\pdfoutput=1
\documentclass[11pt]{article}
\usepackage[final]{acl}
\usepackage{fancyhdr}
\AtBeginDocument{\thispagestyle{fancy}}
\usepackage{times}
\usepackage{latexsym}
\usepackage[T1]{fontenc}
\usepackage[utf8]{inputenc}
\usepackage{microtype}
\usepackage{graphicx}
\usepackage{booktabs}
\usepackage{amsmath}
\usepackage{placeins}
\usepackage{xcolor}
\newcommand{\crnew}[1]{#1}

\graphicspath{{}{judge_ignores_errors/}{metrics_coincide/}{measurement_conditions/}{judge_capping/}{shared/}}

\newcommand{\nProbe}{3{,}600}

\title{Does CoT-Pass@k Really Check the CoT? \\
A Multilingual Mathematical Audit}

\author{Tar{\i}k Tuna Ta{\c{s}}alt{\i}\textsuperscript{1,2} \quad
  Burcu H{\"u}daverdi\textsuperscript{1} \quad
  David Semedo\textsuperscript{2} \\
  \textsuperscript{1}Dokuz Eyl{\"u}l University, {\.I}zmir, T{\"u}rkiye \\
  \textsuperscript{2}NOVA School of Science and Technology, Universidade NOVA de Lisboa, Caparica, Portugal \\
  \texttt{tasaltitariktuna@gmail.com}
}
\begin{document}
\maketitle

\begin{abstract}
Pass@$k$ measures whether a model reaches a correct answer under repeated
sampling, but never how: a lucky guess counts the same as sound reasoning.
CoT-Pass@$k$ was proposed to close that gap, adding an LLM-as-judge that
must assess a solution's reasoning chain before it counts. Its value rests
entirely on one assumption: that the judge catches flawed reasoning. That
assumption has never been tested inside the metric that depends on it, and
never outside English, though the metric's claims concern models used in
many languages. We report the first audit of that verification step, run
under the metric's own protocol on a multilingual suite of five
mathematical benchmarks in English, Turkish and Portuguese, two of them
natively written. We corrupt correct solutions with deterministic edits
that damage the chain and the final answer separately. We observe that
\crnew{all three} judges accept corrupted chains almost as often as clean ones\crnew{.
V4-Flash and Qwen3.6 reject a solution sharply} only when \crnew{its} final answer is wrong\crnew{} and accept a wrong answer more
readily when the chain agrees with it\crnew{; the metric's own judge accepts most wrong
answers as well}. Our study shows that \crnew{chain--answer agreement dominates the
two larger judges' verdicts and that all three fail to reliably detect the
tested reasoning errors}. Consequently the difference Pass@$k$ $-$
CoT-Pass@$k$ averages 19.7 points on an earlier solver generation but only
4.1 on the current one. \crnew{What little remains depends on the token budgets on both sides and on
the generation mode; raising the generation budget moves Pass@$64$ by more than fifty
points while the difference stays at zero.} We close with two checks any
judged reasoning metric should pass before its numbers are read as
evidence about reasoning.
\end{abstract}

\section{Introduction}
\label{sec:intro}

Pass@$k$ measures whether any of $k$ sampled generations answers a question
correctly. It has become the standard instrument for mapping what a model
can reach under repeated sampling, and it anchors the current debate on
whether reinforcement learning with verifiable rewards extends or merely
sharpens a base model's reasoning. But Pass@$k$ never looks at how the
answer was reached: a lucky guess and sound reasoning count the same.

CoT-Pass@$k$ \citep{wen2025reinforcement} was proposed to close exactly
this hole. An LLM judge assesses each correct solution and must approve the
reasoning chain before the solution counts, so the metric promises to
separate models that reason from models that guess. The promise rests
entirely on one assumption: the judge actually assesses the chain.
That assumption is testable; the literature on reasoning-error detection
gives ample reason to doubt it: verifiers are weak at exactly the
step-level judgments the metric needs \citep{jacovi2024reveal, zheng2024processbench}\crnew{, step-level error identification is generally harder than solution-level for every method tested \citep{xia2025reasoneval}}, weaker still on long
chains \citep{he2025deltabench}, and prone
to crediting reasoning that merely looks valid
\citep{wang2025theater}.

In this paper, we audit CoT-Pass@$k$ under its own protocol, comprising the original judge
prompt \crnew{(Figure~\ref{fig:judge-prompt})}, three judgments per solution and the original verification strategies, with controlled error injection: deterministic edits that corrupt a
solution's reasoning chain and its final answer separately
(Section~\ref{sec:method-error-injection}). The audit runs on five
mathematical benchmarks in three languages, including native Turkish and
Portuguese sets (Section~\ref{sec:benchmarks}): the metric is applied to
models used far beyond English, and its judge has only ever been measured
in it.

The audit returns a clear mechanism, and the mechanism predicts the
metric's behaviour. Acceptance barely moves when we corrupt the chain \crnew{under any judge. Under
V4-Flash and Qwen3.6 it collapses} when the
final answer is wrong\crnew{} and rises again when that wrong answer is carried
consistently throughout the chain\crnew{, so chain--answer agreement dominates the
verdicts and the judges fail to reliably detect the tested errors in the steps; the
metric's own judge passes most wrong answers as well}
(Section~\ref{sec:results-error-injection}). It follows
that wherever answers are mostly right, CoT-Pass@$k$ must collapse onto Pass@$k$, and it does: across two solver generations, the average
difference falls from 19.7 points to 4.1
(Section~\ref{sec:results-coincide}). The original
study's solver was a Qwen2.5 base checkpoint, on which the metric does
separate; on the solvers it would be applied to now, what it adds over
Pass@$k$ is close to nothing. Switching the
generation mode reverses which language the same solvers look stronger in,
and raising the token budget moves Pass@$64$ by more than fifty points
while the difference Pass@$64$ $-$ CoT-Pass@$64$ stays at zero at every
step (Section~\ref{sec:results-conditions}).

Our contributions: (i) the first audit of CoT-Pass@$k$'s verification
step, under the metric's own protocol\crnew{, where other error injection tests the judge but not a metric built on it \citep{sun2026enigma, mittal2026c2faith}}; (ii) a controlled, deterministic error-injection
design whose conditions separate chain validity from answer correctness,
including a consistency-preserving wrong-answer condition; (iii) a
multilingual benchmark suite built for this audit\crnew{\footnote{\crnew{Code and the benchmark suite: \url{https://github.com/ttasalti/evalhub}.}}}: translated AIME
2026 \citep{maa2026aime} and native sets from the 2026 Turkish olympiad
\citep{tubitak2026olympiad} and the Portuguese
exams of PHEB \citep{tavares-etal-2026-pheb}. These carry, to our
knowledge, the first CoT-Pass@$k$ \crnew{results} on natively written
Turkish and Portuguese benchmarks; (iv) a two-generation \crnew{comparison} of
the metric's collapse onto Pass@$k$, with question-level uncertainty
quantified; (v) an analysis \crnew{showing that what the metric reports depends on the token budgets, the generation mode and judgments stopped at the max-token limit}; and (vi)
concrete recommendations for reporting and auditing judged reasoning
metrics.

\section{Related Work}
\label{sec:related}

\paragraph{What Pass@$k$ and CoT-Pass@$k$ measure.}
Pass@$k$, adopted from program synthesis \citep{chen2021codex}, is the evidence base for the debate on whether reinforcement learning with
verifiable rewards extends a base model's reasoning or only sharpens its sampling \citep{yue2025does}.
\crnew{Pass@$k$ is a coverage measure that rises with the sample budget
\citep{brown2024monkeys}, which is why its critics add a reliability threshold
\citep{liu2025gpassk, dragoi2025beyondpassk} or replace it with a posterior
success estimate \citep{hariri2026dontpassk}; none of these three metrics inspects the chain.}
CoT-Pass@$k$ \citep{wen2025reinforcement} enters that debate as an
instrument: an LLM judge must approve the reasoning chain before a
solution counts, so the metric should separate reasoning from guessing.
The original study supports its judge by agreement with larger
open-weights judges; Section~\ref{sec:results-error-injection} shows why
agreement is not verification: two judges that both track the final
answer agree for the wrong reason.

\paragraph{\crnew{Turkish and Portuguese benchmarks.}}
\crnew{Turkish has TurkBench \citep{toraman2026turkbench} and Cetvel \citep{er2025cetvel} and Portuguese has MATH-PT \citep{teixeira2026mathpt} and PHEB \citep{tavares-etal-2026-pheb}, but none measures Pass@$k$, and only PHEB grades written solutions, against the official rubrics of open-ended questions our suite does not use; TurkBench draws on TÜB\.ITAK olympiads but predates the 2026 exam. MCLM \citep{son2025mclm} translates AIME 2024 into 55 languages with GPT-4o, Turkish among them, checks only that the answers and equations survive translation, and scores the final answer alone.}

\paragraph{Judge-free \crnew{evaluation}.}
\crnew{Where gold step labels exist, reasoning can be scored without a judge
\citep{uesato2022process, lightman2023letsverify}, and \crnew{ProcessBench \citep{zheng2024processbench} and PRMBench \citep{song2025prmbench} score verifiers against known step errors and find existing process reward models weak at identifying the faulty step.}} These benchmarks
measure judges in isolation; the verdict never leaves the benchmark, so
nothing connects a judge's failure there to the numbers a deployed metric
reports. \citet{sobhani2026mathmist} extends error injection to
multilingual mathematics, Turkish among its languages, though its
injected chains are Bangla and English only;
\citet{zhao2026multilingualcot} corrupt chains across languages to
measure the solver's reliance on its own trace. In concurrent work,
\citet{garcia2026lastword} shows that the effect in such studies tracks
where the answer is stated, not where the computation happens. In all three, a corrupted chain scores a model against a known key\crnew{, the gold answer or, in MathMist, the injected error}; in none does a verdict on a chain become the number a published metric
reports.

\paragraph{Judge-dependent \crnew{evaluation}.}
Where the judge itself is the instrument, its failure modes are documented: preference biases, degradation on long chains, credit for
reasoning that only looks valid \citep{zheng2023judging,
he2025deltabench, wang2025theater}\crnew{, weaker judges accept wrong answers more
readily once a fluent chain is attached \citep{tu-etal-2026-long}, and a chain
need not state the true reason for its answer at all \citep{turpin2023unfaithful}}. Two recent studies come closest to
ours in method. In concurrent work, \citet{sun2026enigma} build
valid-answer, invalid-reasoning solutions and find frontier judges
credit up to half of them as flawless, the verdicts tracking the answer.
\citet{mittal2026c2faith}
inject errors into PRM800K chains and find judges detect that a chain is
wrong far more reliably than where. Both are English-only and stop at the
judge.
\crnew{Our preliminary study \citep{tasalti2026uyik} benchmarks both metrics on
the three AIME 2026 versions and raises, without testing, the question this audit
starts from, whether the judge catches subtle reasoning errors.}
\crnew{Neither study, nor any other we know of,} audits a published judged metric
under the metric's own protocol, follows the judge's verdicts into the
number the metric reports, or ties what remains to \crnew{the token budgets and the generation mode}: the three steps this audit takes, in three languages.

\section{Multilingual Benchmarks}
\label{sec:benchmarks}

The suite holds five benchmarks in three typologically distant
languages: Germanic English, Romance Portuguese, and Turkic Turkish, the
last agglutinative and outside the Indo-European family altogether.
The English core is the AIME 2026
competition set \citep{maa2026aime}: thirty problems with integer answers, on which none of
our solvers reaches the ceiling. Two translated benchmarks carry the
same thirty problems into Portuguese and Turkish: the problems were first
machine-translated, then manually audited by native speakers of each language\crnew{, six for Portuguese and two for Turkish, one of them an author}.
Two native benchmarks complete the suite: the thirty-two first-stage
problems of the 2026 Turkish TÜB\.ITAK mathematics olympiad
\citep{tubitak2026olympiad}, and 166 problems we draw
from PHEB \citep{tavares-etal-2026-pheb}, a multi-subject benchmark of
Portuguese national exams, taking its multiple-choice mathematics
questions with ground-truth answers and converting them to open-ended form\crnew{; two further graduate students, native speakers of Portuguese and independent of the translation audit, checked that each question kept has a single ground-truth answer and can be posed without its options}. All Portuguese
material in the suite is European Portuguese (the variety far less represented in training corpora than Brazilian Portuguese): the original exams, the translation by construction and native audit.
The pairing is deliberate: each non-English language gets one translated
benchmark, which isolates the language while holding the problems fixed,
and one native benchmark, which removes any translation artifact.

The suite spans a wide difficulty range: the AIME sets and the Turkish olympiad
are competition mathematics, the Portuguese exams are high-school level and
nearly saturated by current models. \crnew{Contamination risk is limited for the competition material. The Turkish
olympiad, held on 16 May 2026, postdates the public release of every model we use.
AIME 2026, held in February, postdates the release of Qwen2.5 and R1-distill and
the stated January 2025 training cutoff of Gemma 4; Qwen3.5, Qwen3.6 and V4-Flash publish no training cutoff and appeared only three to eleven weeks after the first exam, little time for its problems to reach their training data. The
Portuguese exams, written between 2006 and 2023, predate every model. AIME problems are mirrored in public corpora within days of release, while \crnew{we found neither the 2026 Turkish olympiad nor the Portuguese exam questions of PHEB on Hugging Face, and the olympiad nowhere but in the organiser's PDF, the} least exposed benchmark of the five.}

\section{Method}
\label{sec:method}

\subsection{Solvers, Judges and Metrics}
\label{sec:method-setup}

\paragraph{Solvers.}
We evaluate two model generations and a second current-generation family.
The earlier generation is Qwen2.5-7B and
Qwen2.5-32B~\citep{qwenteam2024qwen25}, run as base models
and, through their Instruct counterparts, in non-thinking mode. The
current generation is Qwen3.5-4B and Qwen3.5-9B\crnew{~\citep{qwenteam2026qwen35}} in base, non-thinking and
thinking configurations, joined by Gemma-4-E2B-it and
Gemma-4-E4B-it~\citep{gemmateam2026gemma4} in
non-thinking and thinking mode. Base checkpoints run under their
template's default thinking setting, and every solver draws 64 generations per question (16 on the Portuguese exams) at temperature 0.6 and
top-$p$ 0.95, with a 16{,}384-token generation budget unless a section
states otherwise (the error-injection study raises it per benchmark,
Section~\ref{sec:method-error-injection}). Serving templates and the
sample-count choice are given in Appendix~\ref{app:config}.

\paragraph{Judges.}
Two open-weights judges cover the paper: Qwen3.6-35B-A3B\crnew{~\citep{qwenteam2026qwen36}} and
Gemma-4-26B-A4B-it, written Qwen3.6 and Gemma4; the error-injection study
and the cross-generation replication add DeepSeek
V4-Flash~\citep{deepseekai2026v4}, a far larger open-weights model from a
third family that we run through a commercial API. A fourth
judge corroborates the earlier generation \crnew{and scores the
error-injection study as well},
DeepSeek-R1-0528-Qwen3-8B~\citep{deepseekai2025r1, qwenteam2025qwen3},
written R1-distill: it is the verifier of
\citet{wen2025reinforcement} and the reason our earlier generation is the
Qwen2.5 family, so one arm of our comparison reruns the metric on its own
solver and its own judge rather than on a reconstruction of them.

\paragraph{Judging protocol.}
All judges run in
thinking mode at the same temperature 0.6 and top-$p$ 0.95, with a
16{,}384-token judgment budget, and judge every correct solution three
times; Appendix~\ref{app:config} gives the two budget exceptions. A judgment that
reaches its own max-token limit without emitting a verdict counts as neither
an approval nor a rejection; the majority rule, used throughout the paper,
compares approvals against rejections. Solvers are prompted in the language of
the benchmark; their chains, \crnew{like those of the reasoning models studied by \citet{wang2025languagemixing} and \citet{yong2025crosslingual},} are free to switch into English mid-solution\crnew{, and on the non-English benchmarks the Qwen3.5 base checkpoints and every thinking-mode solver write nearly every chain in English or in a mixture, the instruction-tuned solvers in non-thinking mode almost none (Figures~\ref{fig:language-exp2} and~\ref{fig:language-exp3} in Appendices~\ref{app:language} and~\ref{app:mode})}. Verification is therefore
standardised: every solution meets the same instrument, the original
English verification prompt of \citet{wen2025reinforcement}, extended by
two lines (Appendix~\ref{app:prompts}) that name the problem's language
and direct the judge to score the mathematics regardless of the chain's language, keeping the \crnew{judging} directly comparable to the original
study. The other two verification strategies of the original metric are
defined and ablated in Appendix~\ref{app:rules}.

\paragraph{Metrics.}
Pass@$k$ estimates the probability that at least one of $k$ sampled
generations reaches the correct final answer. We compute it with the
standard unbiased estimator $1 - \binom{n-c}{k}/\binom{n}{k}$, averaged
over questions, where $c$ of the $n$ generations of a question are
correct \citep{chen2021codex}. CoT-Pass@$k$ \citep{wen2025reinforcement} counts a
generation as a success only if its final answer is correct \emph{and} the
judge approves its reasoning chain: the same estimator with $c$ replaced
by the number of generations that are both correct and approved. We write
the difference between them out as Pass@$k$ $-$ CoT-Pass@$k$; at $k=n$ it
is the share of questions with a correct solution but no approved one.

\subsection{Error Injection}
\label{sec:method-error-injection}

We test whether the judge actually assesses chains with \emph{error
injection}: we edit solutions so that the reasoning chain and the final
answer are corrupted separately, and we measure what the judge detects.

We use two current-generation solvers, Qwen3.5-4B and Qwen3.5-9B, in
thinking mode, with a generation budget of 65{,}536 tokens for English
AIME, 32{,}768 for the AIME translations and the Turkish olympiad, and
16{,}384 for the Portuguese exams. The original solutions are correctly
answered generations to which all four error types can be applied
(Appendix~\ref{app:config}). From every model$\times$benchmark cell we
randomly select 72 such solutions, giving 720 original solutions. Each is expanded into five variants, the unedited \emph{clean} control and one copy per error type, for a total of $720 \times 5 = \nProbe{}$ variants. The \textbf{intermediate
numeric error} replaces one number drawn at random from between 40\% and
70\% of the solution's length, never one that appears in the question or the answer, \crnew{in the middle of
the solution, where \citet{mittal2026c2faith} confine their injections so that a
judge cannot find them by inspecting the first or last step,} and leaves the
final answer correct. The \textbf{truncation error} deletes
the last 25\% of the reasoning chain and \crnew{re-appends the boxed final answer
on a new line, leaving a visible seam}. With these
two error types the reasoning chain is corrupted but the final answer stays
correct. The \textbf{final-answer error} replaces only the final answer and
leaves the reasoning chain untouched. The \textbf{consistent final-answer
error} replaces every occurrence of the correct final-answer value in the
reasoning chain with the same wrong number, so the chain stays consistent
and supports the wrong final answer. All edits are deterministic string edits (no language model writes the corruptions), and every wrong
number is a near miss of the value it replaces: shifted by one or two, two
digits transposed, or a single digit changed. \crnew{The chain edits adapt the early-answering and adding-mistakes perturbations of \citet{lanham2023measuring}, which truncate a chain or insert one mistaken step into it, to a judge-side test with deterministic edits and no regeneration, and the numeric edits adapt the rule-based numeric perturbations of \citet{singh-etal-2025-exposing}, tightened here to near misses.} The design gives each
error type one job. Models reading a chain are known to \crnew{follow its stated final answer, most strongly at small scale} \citep{garcia2026lastword}, \crnew{and judges to confirm the answer rather than check the steps} \citep{sun2026enigma}; a
rejection under the first two error types can therefore only come from
assessing the chain, and extra acceptance under the consistent error can
only come from rewarding that consistency.

\crnew{Three} thinking-mode judges score every variant, V4-Flash\crnew{, Qwen3.6 and R1-distill}, under
the three-judgment majority rule of
Section~\ref{sec:method-setup}. All comparisons are paired: each judge is
evaluated only on original solutions it judged in all five variants, the same 720 for \crnew{all three} judges. Acceptance rates carry 95\% Wilson intervals\crnew{~\citep{wilson1927probable}}; conditions are compared pairwise with \crnew{an exact McNemar test~\citep{mcnemar1947note}} on the
same solutions, and the paired difference between two conditions is
bounded by a bootstrap over those solutions.

\subsection{Cross-Generation Design}
\label{sec:method-comparison}

To measure whether the difference Pass@$k$ $-$ CoT-Pass@$k$ still opens
in current-generation models,
we compare two solver generations under one fixed \crnew{setting}: the anchor
judge Qwen3.6 with a 16k generation budget, on the four 64-sample
benchmarks, with every curve run to $k=64$, under the majority rule.
Throughout the paper
\emph{generation} refers to the solvers being measured, never to the
judges, which are held fixed across both. Both generations enter as base
checkpoints and, through their instruction-tuned counterparts, in
non-thinking mode (Section~\ref{sec:method-setup}). The two arms hold
different things fixed: in the non-thinking arm the mode is matched
exactly, with thinking explicitly disabled on both sides, while in the
base arm each family's base checkpoint runs under its template's default
thinking setting. The Gemma-4 solvers enter in non-thinking mode only,
and Gemma4 joins the anchor judge on this grid; the two further judges of
Appendix~\ref{app:judges} cover the earlier generation.

A model enters the comparison only if both judges returned verdicts for at
least 30 of its correct solutions in every cell: both modes, all four
benchmarks; a model that misses the threshold anywhere is dropped
entirely, since a partial grid would open the comparison to selection
effects (Appendix~\crnew{\ref{app:judge-matrix}} lists the models this removes).
The Portuguese exams are excluded (they have no non-thinking runs,
and even in the earlier generation the difference there stays between 0 and 3.6 points\crnew{, at that benchmark's $k{=}16$}), and thinking mode is excluded because Qwen2.5 has no
thinking counterpart; thinking-mode solvers are studied in
Section~\ref{sec:results-error-injection}. Intervals on the difference
resample questions within a cell.

\subsection{Token-Budget Measures}
\label{sec:method-budget}

\crnew{Both the solver and the judge} can run out of tokens, and in both cases the
consequence is scored as a failure that the model may not have committed.

On the solver side, a generation still writing at the max-token limit is
cut off before it states a boxed final answer and is graded wrong;
thinking mode makes this common, because the chain consumes the same
budget the answer has to fit in.

On the judge side the same thing happens one level up. A judgment that
spends its whole budget thinking emits no verdict, and under the
majority rule it cannot contribute an
approval: a solution can lose its majority, and be counted against CoT-Pass@$k$, without any judge having rejected it. We count these
stops per judgment and isolate their effect with a counterfactual:
recomputing CoT-Pass@$k$ with stopped judgments counted as approvals.
\crnew{The band between the two curves is an upper bound on how much of the difference
Pass@$k$ $-$ CoT-Pass@$k$ the judge's token budget, rather than its verdicts,
produced, since some of the stopped judgments would have ended in rejection had
they finished.}

\crnew{We also run a budget ladder that raises the generation budget itself (Section~\ref{sec:results-conditions}).}

\section{Results}
\label{sec:results}

\subsection{The Judges Miss the Injected Reasoning Errors}
\label{sec:results-error-injection}

\begin{figure*}[t]
  \centering
  \includegraphics[width=\textwidth]{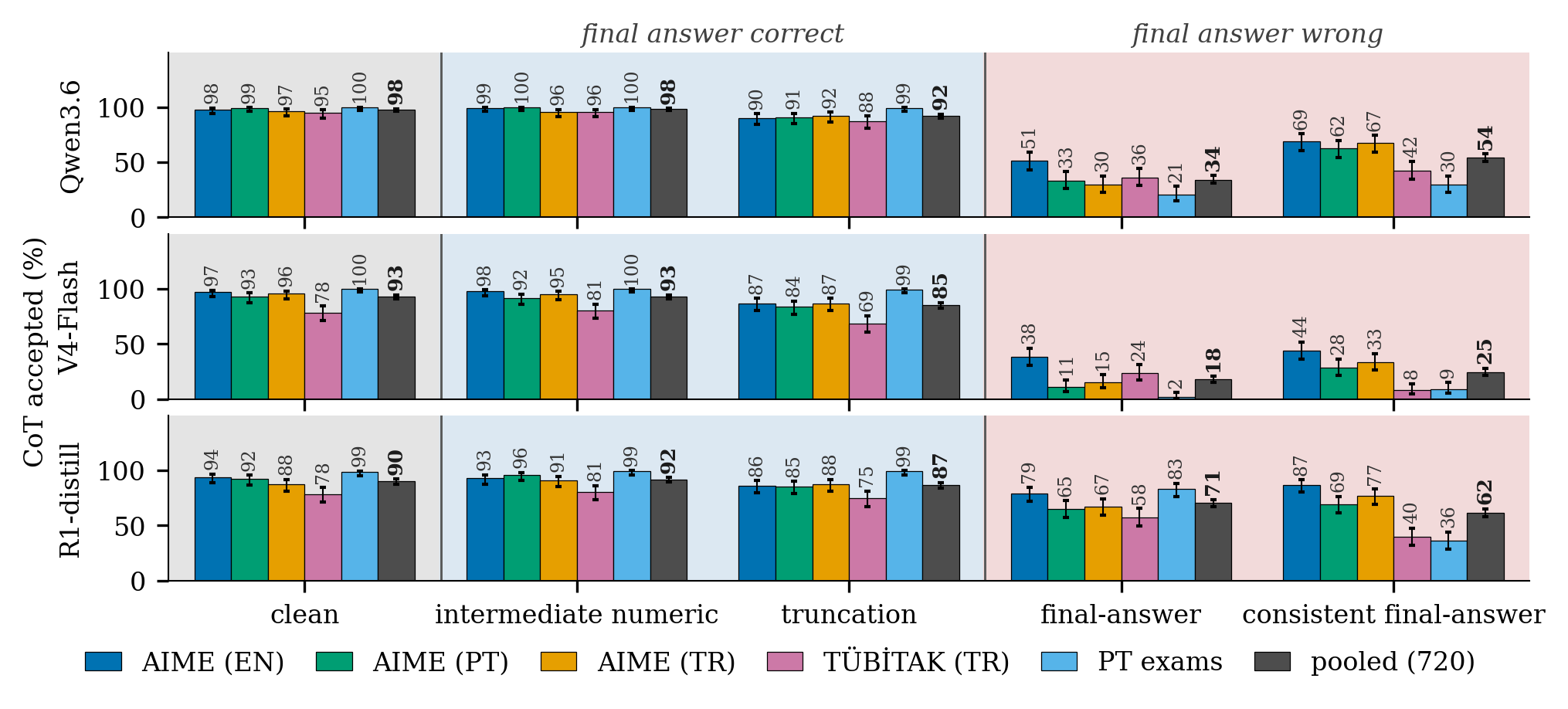}
  \caption{\crnew{CoT acceptance under injected errors per benchmark and judge (rows),
  majority-correct strategy of Section~\ref{sec:method-setup}, 95\% Wilson intervals;
  the dark grey bar is the rate pooled over the five benchmarks. Every judge assesses
  the same 720 original solutions, 144 per benchmark and condition. R1-distill is the
  judge the metric specifies.}}
  \label{fig:error-injection}
\end{figure*}

Figure~\ref{fig:error-injection} shows acceptance for the five variants
under majority-correct. \crnew{All three} judges accept the unedited clean control at
high rates, 93\% for V4-Flash\crnew{,} 98\% for Qwen3.6 \crnew{and 90\% for R1-distill}.

The two error types that keep the final answer correct barely move them.
Under the intermediate numeric error \crnew{all three} judges stay at their
clean rates, 93\%\crnew{,} 98\% \crnew{and 92\%}, although every edited chain now \crnew{carries a number that the steps around it no longer support}: a paired McNemar test over
the same 720 solutions finds no evidence of a shift from the clean
control ($p = 1$\crnew{, $p = 0.55$ and $p = 0.23$}), and the paired difference is bounded
within \crnew{1.5} points in either direction \crnew{for V4-Flash and Qwen3.6 and within 5 for
R1-distill} (95\% bootstrap interval\crnew{s}). The
truncation error costs 8\crnew{,} 6 \crnew{and 3} points.

Corrupting the final answer gives the opposite picture. The final-answer
error leaves the chain untouched and replaces only the number at the end,
and acceptance collapses at once, to 18\% for V4-Flash and 34\% for
Qwen3.6\crnew{, and only to 71\% for R1-distill}.
\crnew{V4-Flash and Qwen3.6} can reject, sharply and in agreement with each other, but they did so only when the final answer was wrong. \crnew{The metric's own judge barely rejects even there, leaving 71\% of the wrong answers accepted.} \crnew{Changing a number inside the chain costs no judge a measurable share of its acceptance; changing only the number at the end costs 75 and 64 points for V4-Flash and Qwen3.6 and 20 for R1-distill.}

The consistent final-answer error shows why. It corrupts strictly more of
the reasoning than the final-answer error, since the wrong value now runs
throughout the chain, yet acceptance rises instead of falling: to 25\% for
V4-Flash and to 54\% for Qwen3.6, which accepts the majority of these
variants. A judge that verified the mathematics would order the two the
other way round. Both conditions are built from the same 720 solutions, so
the ordering can be tested pairwise: McNemar's test rejects equality for
\crnew{V4-Flash and Qwen3.6} (\crnew{$p = 2.4\times10^{-3}$ and $p = 2.6\times10^{-14}$}), with the
discordant pairs running the same way for each.
\crnew{R1-distill does not show the rise, accepting 62\% of the consistent
variants against 71\% of the plain ones ($p = 1.7\times10^{-4}$;
Appendix~\ref{app:rules}).}
\textbf{\crnew{None of the three judges reliably detects the tested errors in the
steps, and for V4-Flash and Qwen3.6 chain--answer agreement dominates the
verdicts.}}

\crnew{The verification strategy changes the rates, not the findings (Table~\ref{tab:error-rules} in Appendix~\ref{app:rules}), and the size of the edit matters only for the consistent wrong answer; the intermediate error is missed at every size (Figure~\ref{fig:edit-size}).}

\crnew{The pattern repeats on every benchmark (Figure~\ref{fig:error-injection};
Appendix~\ref{app:rules} gives the rates per benchmark). Under every judge the
intermediate numeric error stays within four points of the clean control on all
five, and the translated AIME sets behave like the English original. A wrong
final answer costs V4-Flash and Qwen3.6 47 to 98 points everywhere, most on the
short Portuguese exams, and R1-distill 15 to 27, and the Turkish olympiad is where
the judges doubt correct chains too.}

\subsection{CoT-Pass@k Collapses onto Pass@k in Current Solvers}
\label{sec:results-coincide}

\begin{figure}[t]
  \centering
  \includegraphics[width=0.80\columnwidth]{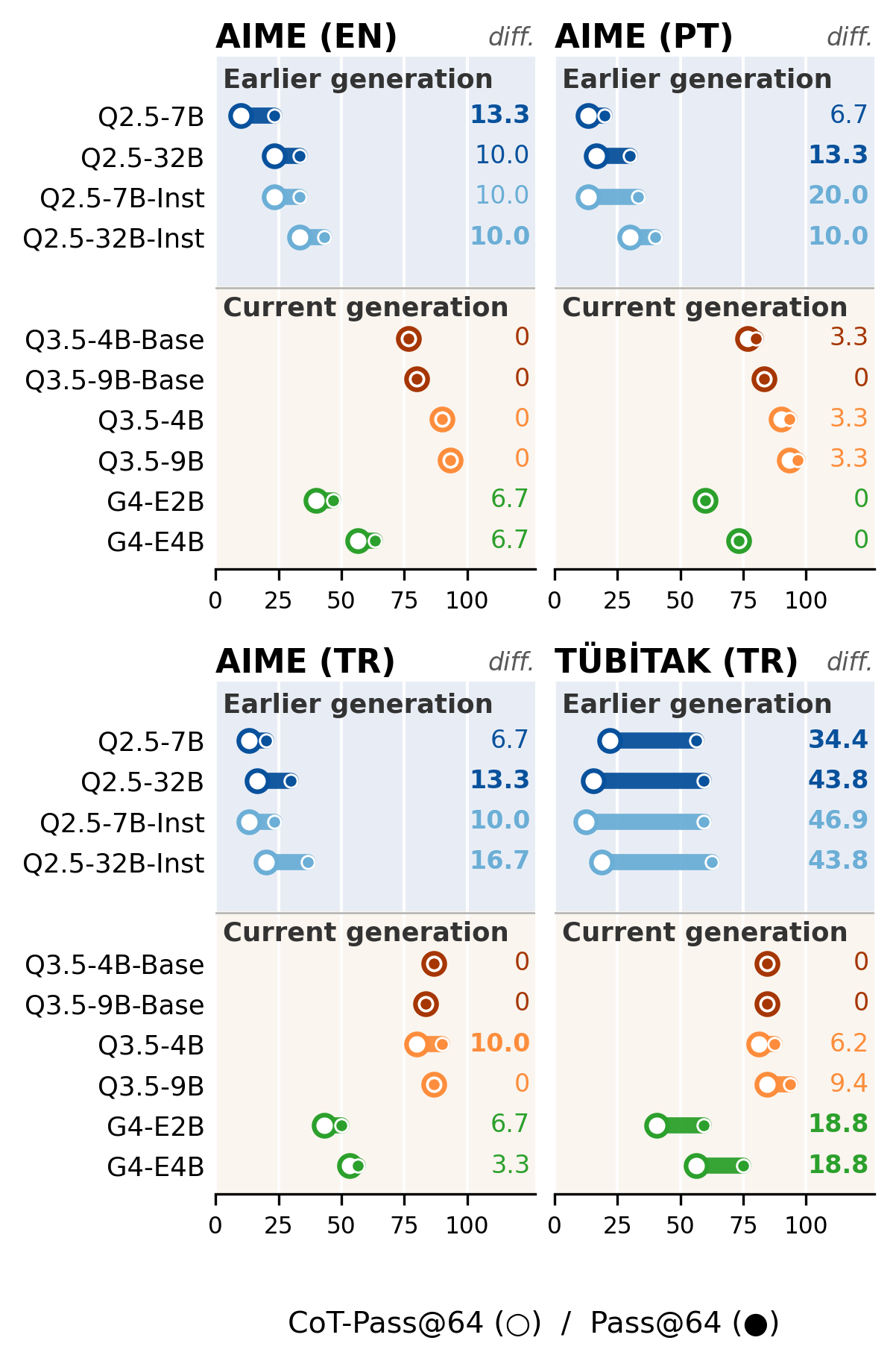}
  \caption{CoT-Pass@64 (open) and Pass@64 (filled) per solver and
  64-sample benchmark under the anchor judge Qwen3.6 at 16k; row-end numbers give the
  difference. Q2.5/Q3.5 = Qwen2.5/Qwen3.5, G4 = Gemma-4-it.}
  \label{fig:gap-endpoints}
\end{figure}

Section~\ref{sec:results-error-injection} predicts where CoT-Pass@$k$
should collapse onto Pass@$k$: the judge \crnew{mostly} approves a chain that agrees with its final
answer, so the difference Pass@$k$ $-$ CoT-Pass@$k$ can come only from
correct answers carrying chains the judge rejects, and should vanish
where those are rare. Figure~\ref{fig:gap-endpoints} confirms the
prediction. Among
the earlier-generation solvers (Qwen2.5), the difference Pass@$64$ $-$
CoT-Pass@$64$ is large for every model on every benchmark: 6.7--20 points
on the AIME benchmarks and 34--47 points on the Turkish olympiad, for both
base and non-thinking models.

Among the current-generation solvers (Qwen3.5 and Gemma-4) the same
\crnew{judge and budget find} almost nothing. Qwen3.5-9B-Base ends at exactly zero on all four
benchmarks, Qwen3.5-4B-Base on three of the four, and no Qwen3.5 model
exceeds 10 points anywhere in it; the Gemma-4 models stay at or below 6.7 on the
AIME benchmarks and reach 18.8 only on the Turkish olympiad, still well below
the smallest earlier-generation value there, 34.4. Averaged over the
four benchmarks, the difference is 19.7 points (95\%~CI $[16.4, 23.0]$)
in the earlier generation against 4.1 ($[2.7, 5.5]$) in the current one (per solver, 15.6--22.1 against at most 8.2), and the contrast
itself is 15.6 ($[12.0, 19.3]$). The intervals are bootstrap intervals \crnew{over questions, the sampled unit in the framework of \citet{miller2024errorbars}}; Figure~\ref{fig:bootstrap} in
Appendix~\crnew{\ref{app:k-ladder}} gives one for every solver. \textbf{\crnew{On current-generation solvers, CoT-Pass@$k$ stays within a few points of Pass@$k$, and the judge changes almost nothing of what Pass@$k$ already reports.}}

Nor does this depend on the anchor judge. Four different judge models,
among them the verifier of the original study, reproduce the earlier
generation's average difference within two points of each other; Gemma4,
the only other judge that covers both generations in full, reproduces
the split between them (Tables~\ref{tab:judge-matrix-earlier}
and~\ref{tab:judge-matrix-current} in Appendix~\crnew{\ref{app:judge-matrix}}, per
judge and per model). Under
both judges that span the generations, the collapse is a property of the
metric, not of the judge we chose to anchor it on.

The divergence is not present from the start: at $k=1$ the generations
are indistinguishable, 2.7 points against 2.0, and
Figure~\ref{fig:k-ladder} in Appendix~\crnew{\ref{app:k-ladder}} follows them
apart to 19.7 against 4.1. What
separates them is what repeated sampling adds: in the earlier generation
each new correct answer had a growing chance of carrying a chain the
judge would reject; in the current one it does not.

\crnew{Per generation the verdicts do change. Averaged over the four benchmarks,
the anchor judge approves 61.6\% of the earlier generation's answer-correct
generations and 94.8\% of the current generation's, and Gemma4 62.7\% and
93.2\% (Figure~\ref{fig:acceptance} in Appendix~\ref{app:acceptance}). Two things therefore
shrink the difference. The judges reject far fewer of the current generation's
correct chains, and the rejections that remain rarely reach CoT-Pass@$64$, because
a question keeps its CoT-Pass@$64$ as long as one of its $c$ correct generations
is approved, and the current generation's solved questions typically hold dozens
(Appendix~\ref{app:saturation}). Comparing the generations at the same $c$ leaves the gap in place (Figure~\ref{fig:saturation}).}

The collapse holds on all four benchmarks, across three languages, and at
every $k$ we measure, with the earlier-minus-current contrast carrying a
95\% interval that excludes zero on each benchmark separately. Three
readings short of a generational change do not survive the data. \crnew{Higher accuracy alone does not produce it. At an identical
Pass@$64$ of 59.4 on the Turkish olympiad, the two earlier-generation
solvers lose 43.8 and 46.9 points where the current one loses 18.8
(Figure~\ref{fig:gap-endpoints}), and the acceptance rates above separate the
generations at every level of accuracy (Figure~\ref{fig:saturation}).} It
is not model size: scaling Qwen2.5 from 7B to 32B leaves the difference
as large, 34.4 against 43.8, while at the closest matched scale, Qwen2.5-7B against the larger Qwen3.5-9B, the difference falls from
15.6 to zero in the base arm and from 22.1 to 3.3 in the non-thinking
one. And it is not the mode mix: the mode-matched arm alone, Qwen2.5-Instruct against Qwen3.5 non-thinking, shows the collapse
from 10--47 points down to at most 10. \crnew{The comparison remains observational, between model families rather
than a controlled intervention, and isolates none of what produces the change.}

\subsection{Raising the Budget Moves the Scores, Not the Difference}
\label{sec:results-conditions}

The residual that Section~\ref{sec:results-coincide} leaves behind is small
under the anchor judge. Both sides of the metric
spend token budget, the judge to take in the chain and still reason, the
solver to finish the chain and still answer, and both move the numbers
that get reported (Table~\ref{tab:conditions} \crnew{in Appendix~\ref{app:stops}}, at the shared 16k budget).

The judge's side first. The two judges stop at their own max-token limit
at very different rates. Judging the Qwen3.5 solutions on the four
competition benchmarks, Gemma4 stops on 33 to 54\% of its judgments;
judging the Gemma-4 solutions on the same benchmarks, on 2 to 14\%.
Qwen3.6 stays at or below 18\% throughout, and most of its stops come
after a verdict has already been written, while Gemma4's stops \crnew{rarely}
contain one. A stopped judgment cannot contribute an approval, so
it can only widen the difference
(Section~\ref{sec:method-budget}). And the two
judges do report different differences on the same solutions: averaged
over these four benchmarks at $k=64$, 5.7 points under Gemma4 against
1.4 under the anchor. The counterfactual of
Section~\ref{sec:method-budget} measures how much of that is budget:
counting Gemma4's stopped judgments as approvals removes 92.7\% of its
5.7\crnew{, an upper bound on the budget's share}, against 47.9\% in non-thinking mode, where the stops are rarer and
the two judges nearly agree to begin with, 5.9 against 5.8
(Figure~\ref{fig:capping} in Appendix~\ref{app:capping}). Under Gemma4, \crnew{the judge's token budget can account for most of what the
metric reports in thinking mode}; under Qwen3.6, on the very same
solutions, there is almost nothing left to report. The collapse itself
shows under either judge (Section~\ref{sec:results-coincide}); what
remains on top of it can be an artifact of the judge one happens to pick.

The solver's side mirrors it one level down, and the cleanest view
changes nothing but the generation mode. On the same thirty problems,
enabling thinking raises the share of English generations that stop at
the max-token limit from 27.9\% to 86.7\% (the chains grow, the budget does not) while the Turkish rate rises only to 39.9\%
(Table~\ref{tab:language} in Appendix~\ref{app:mode}). Pass@$64$ inverts
with it, English against Turkish: 91.7 to 88.3 in non-thinking mode, 40.0
to 85.0 with thinking on. Turning on the chain of thought that CoT-Pass@$k$ exists
to check reverses which of the two languages these models look stronger
in. We therefore read language differences throughout as \crnew{effects of the
generation mode and the token budget}, not statements about ability in a language\crnew{; the mode even sets the
language the chain is written in (Figure~\ref{fig:language-exp3} in
Appendix~\ref{app:mode})}.

The benchmarks line up the same way: the Portuguese exams, the easiest
of the five and the least bound by the max-token limit on either side
(Table~\ref{tab:conditions}), carry the smallest difference, 0.60 points at most\crnew{, at that benchmark's
$k{=}16$}. Where solutions fit the budgets, there is nothing left for
the metric to report. The Turkish olympiad marks the one residual the
budget reading does not carry: the largest difference the anchor judge
reports here, 9.38 points under the Gemma-4-E4B solver, sits where
essentially nothing stops at the max-token limit, not one of that
solver's generations, and 0.1\% of the anchor judge's. Whatever it
tracks, it is not the budget; we return to it in
Section~\ref{sec:discussion}.

So far the budget has been held fixed and everything else varied; moving
it turns the reading into an intervention. Table~\ref{tab:ladder} in
Appendix~\ref{app:ladder} compares the two solvers of
Section~\ref{sec:method-error-injection} at a low and a high generation
budget: the same problems, more room to write. The high budget is 32k,
except 64k on English AIME, whose chains still ran into the max-token
limit at 32k.

The solver's side of Table~\ref{tab:ladder} responds exactly as
Table~\ref{tab:conditions} predicts: the share of generations stopping
at the max-token limit collapses, from 84--90\% to 9--18\% on English AIME, and accuracy climbs with it,
Pass@$1$ from 10.5 and 16.2 to 75.4 and 84.5 there, Pass@$64$ from 40.0 to
96.7 there and by 3 to 17 points on the other benchmarks. What CoT-Pass@$64$
adds on top of Pass@$64$ never moves at all. The ladder's solutions are
judged as everywhere else, by the anchor judge at a budget that rises
with the solver's, and the difference Pass@$64$ $-$ CoT-Pass@$64$ is
exactly zero in all sixteen benchmark $\times$ solver $\times$ budget
cells.
\crnew{The anchor judge still rejects some solutions, but no question ever loses all of its correct ones.}
\textbf{CoT-Pass@$64$ follows Pass@$64$ point for point up the ladder:
whatever the budget produces, the metric certifies.}
The
ladder is reported under the anchor judge alone \crnew{(Appendix~\ref{app:ladder})}.

\section{Discussion and Conclusion}
\label{sec:discussion}

\paragraph{What the metric measures.}
A judgment that
approves a corrupted chain, rejects it once its final answer is wrong,
and approves it again once that wrong value runs throughout the chain is a
judgment \crnew{in which agreement between the chain and the final answer dominates
and the tested errors in the mathematics count for little}. \crnew{The metric's own judge fits only the first clause, since it approves
most wrong answers too.}
That agreement is nearly free for a solver whose answers are usually
right, which is why, on current-generation solvers, the metric returns
what Pass@$k$ already returns. What it reports is thus a property of the
solver--judge pair: the same judge filters the two generations
differently, and the same solvers score differently under a judge that
stops at the max-token limit.

\paragraph{Two checks.}
The question we put to CoT-Pass@$k$ is the one \crnew{work on construct validity} puts
to any instrument, whether it captures the construct it names \citep{bean2025construct}, and our results turn it into two checks
that a judged reasoning metric should pass before its numbers are read as
evidence about reasoning. \textbf{(i) Publish how the judge responds to injected errors.}
Give it solutions whose chains are corrupted while their final answers
stay correct; if approval does not move, nothing the metric reports
separates a sound chain from a corrupted one. The check is run once, by whoever
proposes the judge, on \crnew{several} hundred solutions\crnew{; in our runs the whole error-injection panel took about a sixth of the judge tokens the anchor judge spent on the solutions of Section~\ref{sec:results-coincide} (Appendix~\ref{app:cost})}. \textbf{(ii) Report the metric
beside the judge-free metric it wraps, under more than one \crnew{token budget and generation mode}.} The difference between the two is all the judge adds, and
publishing it per solver and per benchmark, at more than one token
budget, costs a column.

\paragraph{Where the metric still separates.}
None of this makes chain-level verification a bad idea; it makes an
unaudited judge a bad instrument. The metric does separate the earlier
generation, and \crnew{the original study itself observes that the separation it reports shrinks on benchmarks the base model already solves} \citep{wen2025reinforcement}, but harder problems
do not restore its meaning, since the judges do not react to corrupted
chains there either. What the metric cannot carry is the debate it
entered: on current-generation solvers a conclusion drawn from it
inherits the Pass@$k$ evidence it was meant to go beyond, and whether
verifiable rewards extend reasoning has to be settled with \crnew{instruments that measure the chain directly, such as the causal-importance and sufficiency metrics of \citet{yu2026outcome}, who find that verifiable rewards raise accuracy without reliably making the chain causally important or sufficient}.

\paragraph{\crnew{What the languages add.}}
\crnew{The verifier failure is not English-specific in the tested settings. On the
same thirty AIME problems in three languages the judges miss the injected errors,
reject wrong answers and approve clean chains at rates at most 6.3 points below the English original's, and sometimes above them (Section~\ref{sec:results-error-injection}), and averaged over solvers the translations are approved within 3 points of the original or above it (Appendix~\ref{app:acceptance}), so we find no sign that Turkish or Portuguese makes
verification harder for these judges. What moves the rates is the native benchmarks, the Turkish olympiad on the correct chains and the Portuguese exams on the wrong answers, and the olympiad also carries the one difference the budget reading does not explain
(Section~\ref{sec:results-conditions}), for which the Limitations name the candidates. The
language a chain is written in follows the solver rather than the benchmark, and its effect on the verdicts is small and inconsistent: within 3 points for the Qwen3.5 base checkpoints in every cell, 13 points in favour of English for one earlier-generation solver on the olympiad, and in favour of mixed chains under V4-Flash and R1-distill there (Figures~\ref{fig:code-switching} and~\ref{fig:language-exp1}, Appendix~\ref{app:language}).}

\paragraph{Conclusion.}
An instrument that promises to assess reasoning has to be audited the way
any instrument is, against inputs whose correct verdict is known by construction\crnew{, as BLEU once was against constructed variants that human judges would rank far lower \citep{callisonburch2006reevaluating}}. Under that audit CoT-Pass@$k$'s verification step \crnew{fails to
reliably detect the tested errors and, under V4-Flash and Qwen3.6, its verdicts
are dominated by chain--answer agreement}, what the metric adds over Pass@$k$ has collapsed on
current-generation solvers, and what remains of it \crnew{depends largely on the token
budgets on both sides and on judgments stopped at the max-token limit}.

\section*{Limitations}

\paragraph{The injection is deterministic, and deliberately so.}
String edits keep the correct verdict known by construction: no second
model has to be trusted to corrupt a chain, and every variant is
reproducible. The cost is realism, and a lenient reader might treat an
edited step as a slip rather than broken reasoning. That reading cannot
explain the consistent final-answer error, which corrupts strictly more of
the chain than the plain one yet is accepted more often\crnew{ by V4-Flash and Qwen3.6}; and where our
edits are too mild, the bias understates rather than overstates how little
the judges assess.

\paragraph{\crnew{What the edits cover.}}
\crnew{Truncation leaves a visible seam, so the 3 to 8 points it costs are an upper bound on what a chain that stops short costs these judges. The injected solutions come from two Qwen3.5 thinking-mode solvers, so the judges are audited on one family's style of chain, and on the non-English benchmarks these chains are written in English or in a mixture rather than in the benchmark's language (Appendix~\ref{app:language}), so outside English the judges are audited on non-English problems with English or mixed reasoning. We did not check by hand whether each edited number carries into the steps that follow; an edit in an auxiliary line corrupts less of the mathematics than a broken derivation, and the consistent final-answer condition, whose corruption needs no such check, carries the ordering result.}

\paragraph{Scale is bounded on both sides\crnew{, solver and judge}.}
Our current-generation solvers reach 9B parameters, so the collapse is not
directly verified on larger current models; within the earlier generation,
where we do scale, size does not produce it, since 7B to 32B leaves the
difference as large. On the judge side, the audit rests on three judges
larger \crnew{in total parameters} than the one the metric specifies (Qwen3.6, Gemma4 and DeepSeek
V4-Flash) and on that original 8B judge itself \citep{wen2025reinforcement}, \crnew{which in our error-injection study misses the chain-level errors as they do but rejects a wrong final answer far less often}; it does not reach the flagship tier of any
lab. \crnew{The judges differ in how much they accept, and the weakest one also differs in what it accepts. None of them reacts to the intermediate numeric error. V4-Flash and Qwen3.6 accept the consistent wrong answer more often than the plain one, R1-distill does not, and it accepts 71\% of the plain wrong answers outright. We read the original judge's behaviour off its
acceptance rates and did not audit the reasoning behind its verdicts, so
whether it accepts on the form of a solution rather than on its content
remains open.}

\paragraph{The residual is observed, not explained.}
The collapse is least complete on the Turkish olympiad: against
earlier-generation differences of 34--47 points there, the
current-generation solvers fall far under the anchor judge, the Qwen3.5
solvers to at most 9.4 points and the Gemma-4 solvers to 18.8
(Figure~\ref{fig:gap-endpoints}), but not to zero. That benchmark is at
once natively written, the least publicly exposed of the five, and at
olympiad level, and our suite cannot separate these: the Portuguese exams
are natively written too, but they are high-school level and nearly
saturated, so they cannot isolate the effect of native language. \crnew{The
multilingual reach is three languages in one domain, so the findings are not
English-specific in the tested settings, and we claim nothing beyond those
settings.} The audit
is also mathematical throughout, and whether the same behaviour holds in
code or in domains with no single verifiable answer, our data cannot say.

\section*{Ethics Statement}

\crnew{This work evaluates publicly released language models on published mathematical
exams, used for research evaluation only; no personal data and no human subjects are
involved. The translations were audited by an author and volunteer native speakers,
and the corrupted solutions exist only as test inputs to the judges. The audit concerns
a published metric, not its authors: we run it under its own prompt and protocol and
report where it still separates models. Generation and judging ran largely on H200
GPUs, with part of the judging through commercial APIs. The examination material
belongs to its original sources; the AIME problems are credited to the MAA AMC and used
for non-commercial research, as are our translations of them. We used AI assistants for code, figure scripts,
literature checks and language editing.}

\section*{Acknowledgments}

This work was supported by the AMALIA project under Measure RE-C05-i08 of the
Portuguese national Programa de Recupera{\c{c}}{\~a}o e Resili{\^e}ncia. We also
acknowledge the support of the NOVA LINCS project (UID/04516/2025).

\bibliography{custom}

\appendix

\setcounter{topnumber}{4}\setcounter{bottomnumber}{3}\setcounter{totalnumber}{6}
\renewcommand{\topfraction}{0.95}\renewcommand{\bottomfraction}{0.8}
\renewcommand{\textfraction}{0.05}\renewcommand{\floatpagefraction}{0.8}
\setcounter{dbltopnumber}{2}
\renewcommand{\dbltopfraction}{0.95}\renewcommand{\dblfloatpagefraction}{0.9}

\section{Experimental Setup}
\label{app:setup}

\subsection{Prompts}
\label{app:prompts}

\paragraph{Solver prompts.}
Every solver receives the problem statement followed by one
instruction sentence in the language of the benchmark:

\begin{quote}\small\raggedright
\textbf{EN:} Let's think step by step and output the final answer within
\texttt{\textbackslash boxed\{\}}.\\[2pt]
\textbf{TR:} Adım adım düşün ve nihai cevabı
\texttt{\textbackslash boxed\{\}} içerisinde ver.\\[2pt]
\textbf{PT:} Vamos pensar passo a passo e apresentar a resposta final
dentro de \texttt{\textbackslash boxed\{\}}.
\end{quote}

The Turkish line serves the AIME translation and the Turkish olympiad; the
Portuguese line serves the AIME translation and the Portuguese exams.

\paragraph{Judge prompt.}
Figure~\ref{fig:judge-prompt} gives the English judge prompt in full; it
reproduces the verification prompt of
\citet{wen2025reinforcement}. The judge fills \texttt{\{question\}} and
\texttt{\{solution\}} and must end with
\texttt{\textbackslash boxed\{yes\}} or \texttt{\textbackslash boxed\{no\}}.

\begin{figure*}[!t]
\small\ttfamily\raggedright\hyphenpenalty=10000\exhyphenpenalty=10000
\noindent\rule{\textwidth}{0.4pt}\\[2pt]
You are an expert in mathematics and logical reasoning. Your task is to
evaluate the correctness of a solution to a given math problem, with a
**strong emphasis on the reasoning process**, not just the final answer.\\
Below is the **Problem** and the **Solution (Provided by another AI
model)**:\\
---\\
**Problem**:\\
\{question\}\\
**Solution (Provided by another AI model)**:\\
\{solution\}\\
---\\
Please perform the following tasks:\\
1. **Analyze the solution step-by-step**, paying close attention to: -
Computational accuracy - Logical consistency - Conceptual understanding -
Whether the reasoning is valid and complete\\
2. **Identify any issues or errors in the reasoning**, even if the final
answer is correct. Classify them into the following categories (if
applicable): - **Calculation Error**: Mistakes in arithmetic, algebraic
manipulation, or numerical computation. - **Logical Error**: Invalid
reasoning, flawed logic, or incorrect inference. - **Conceptual Error**:
Misunderstanding or misuse of mathematical concepts or definitions. -
**Omission / Incompleteness**: Missing steps, incomplete justification, or
not addressing all parts of the question. - **Other**: Any other type of
error that does not fit into the above categories.\\
3. **Provide a final judgment** on whether the solution is logically sound
and free of errors in reasoning.\\
Please format your response as follows:\\
---\\
**Issues Identified:**\\
- [Issue 1]: [Classification] - [Brief explanation] - [Issue 2]:
[Classification] - [Brief explanation] - ...\\
Let's think step by step and output your final judgment within
\textbackslash boxed\{\}\\
\textbackslash boxed\{yes\} or \textbackslash boxed\{no\}\\[2pt]
\noindent\rule{\textwidth}{0.4pt}
\rmfamily\normalsize
\caption{The English judge prompt, reproduced from
\citet{wen2025reinforcement}. \texttt{\{question\}} and
\texttt{\{solution\}} are filled per judgment; the Turkish and Portuguese
variants differ from it by the two additions given in Appendix~\ref{app:prompts}.}
\label{fig:judge-prompt}
\end{figure*}

\paragraph{Multilingual variants.}
The Turkish and Portuguese judge prompts are identical to the English one
except for exactly two additions. First, the opening sentence tags the
problem language: ``a given math problem \textbf{(written in Turkish)}''
(respectively \textbf{Portuguese}). Second, a fourth task is appended:

\begin{quote}\small\ttfamily\raggedright
4. **Language Consideration**: Ignore whether the solution is provided in
Turkish, English, or a combination of both (language switching). Focus
exclusively on mathematical and logical correctness, disregarding the
language used in the evaluation.
\end{quote}

Everything else (the task list, the error taxonomy, the output format, and the verdict convention) is unchanged from
\citet{wen2025reinforcement}, so the multilingual judgments are the same
\crnew{instrument} as the original English one up to these two additions.

\subsection{Configuration}
\label{app:config}

\paragraph{Serving and sampling.}
All models are served with their family's official chat templates: for the
Qwen3.5 instruct models thinking is explicitly disabled or enabled, while
base checkpoints run under their template's default thinking setting;
sampling parameters are set explicitly for every run. Every solver draws 64
generations per question, and 16 on the Portuguese exams, whose
near-saturation makes larger samples uninformative and whose 166 questions
keep the total sample count comparable to the other benchmarks. Our
pipeline extends an open-source evaluation harness with these judging
stages.

\paragraph{Judge budgets.}
The judgment budget is 16{,}384 tokens throughout, with two exceptions.
V4-Flash judges at 20{,}480 tokens wherever it appears, and in the
error-injection study of Section~\ref{sec:method-error-injection} \crnew{all three}
judges share that setting; the budget ladder of
Section~\ref{sec:method-budget} raises the anchor judge's budget in step with the solver's.

\paragraph{\crnew{Judge version.}}
\crnew{V4-Flash is the April 2026 preview build of DeepSeek-V4-Flash, queried through the DeepSeek API before 31 July 2026, when the same API name moved to re-trained builds; the preview weights remain published.}

\paragraph{\crnew{Judging cost.}}
\label{app:cost}
\crnew{Pass@$k$ needs only the solver's generations, while CoT-Pass@$k$ also sends every answer-correct generation to the judge three times. On the judge matrix of Section~\ref{sec:results-coincide} the ten solvers produced 78{,}080 generations with 416 million output tokens, of which 20{,}059 were answer-correct. Qwen3.6 and Gemma4 judged all of these, 60{,}177 judgments each, and V4-Flash and R1-distill, which cover the earlier generation only, judged its 2{,}180 answer-correct generations, 6{,}540 judgments each, for 133{,}434 judgments and 797 million judge output tokens in all. A judgment averages 3.5 thousand output tokens under V4-Flash and 7.9 thousand under R1-distill, so judging one correct solution costs 10 to 24 thousand tokens, and in the same cells Qwen3.6 and Gemma4 emit 0.82 and 0.92 tokens for every solver token, which nearly doubles the output of a Pass@$k$ evaluation. The error-injection study adds 32{,}400 judgments, 720 solutions in five variants, three judgments each, under three judges. The jobs in our cluster's accounting add up to about 620 H200 GPU-hours of generation and open-weights judging, exploratory runs included; this is a lower bound, since runs outside that record, V4-Flash, which ran through the DeepSeek API, and the Qwen3.6 judgments of two error-injection cells, served through a commercial API when the cluster was unavailable, are not in it.}

\paragraph{Selecting solutions for error injection.}
The original solutions of Section~\ref{sec:method-error-injection} are
drawn from correctly answered generations that stopped before the
max-token limit and are longer than 600 tokens, keeping only solutions to
which all four error types can be applied.

\section{CoT Verification Strategies}
\label{app:rules}

\begin{table}[t]
\centering
\footnotesize
\begin{tabular}{lccc}
\toprule
Variant & any & majority & all \\
\midrule
\multicolumn{4}{l}{\textit{V4-Flash}} \\
Clean & 96.2 & 92.9 & 89.2 \\
\cmidrule(lr){1-4}
Intermediate numeric error & 96.1 & 93.1 & 90.1 \\
Truncation error & 89.3 & 85.1 & 78.8 \\
Final-answer error & 47.2 & 18.1 & 6.4 \\
Consistent final-answer error & 49.0 & 24.6 & 12.4 \\
\midrule
\multicolumn{4}{l}{\textit{Qwen3.6}} \\
Clean & 99.4 & 97.8 & 81.2 \\
\cmidrule(lr){1-4}
Intermediate numeric error & 99.2 & 98.2 & 81.0 \\
Truncation error & 93.9 & 92.1 & 69.2 \\
Final-answer error & 67.2 & 34.3 & 7.8 \\
Consistent final-answer error & 75.6 & 54.2 & 27.8 \\
\midrule
\multicolumn{4}{l}{\crnew{\textit{R1-distill}}} \\
\crnew{Clean} & \crnew{96.9} & \crnew{90.1} & \crnew{36.4} \\
\cmidrule(lr){1-4}
\crnew{Intermediate numeric error} & \crnew{96.0} & \crnew{91.9} & \crnew{42.2} \\
\crnew{Truncation error} & \crnew{92.1} & \crnew{86.7} & \crnew{67.4} \\
\crnew{Final-answer error} & \crnew{91.9} & \crnew{70.6} & \crnew{28.3} \\
\crnew{Consistent final-answer error} & \crnew{80.3} & \crnew{61.8} & \crnew{23.1} \\
\bottomrule
\end{tabular}
\caption{Acceptance rates (\%) under the three verification strategies of the original CoT-Pass@$k$ judge: a variant is accepted when at least one of its three judge generations approves it (any-correct), when more approve than reject (majority-correct), or when all three do (all-correct). The majority-correct column is the one plotted in Figure~\ref{fig:error-injection}. 720 original solutions per error type for \crnew{each judge; R1-distill is the judge the metric specifies}.}
\label{tab:error-rules}
\end{table}

The original CoT-Pass@$k$ judge turns the three judgments of a solution into
one verdict with a \emph{verification strategy}: \emph{any}-correct accepts
a variant if at least one of its three judgments approves it,
\emph{majority}-correct if more of them approve than reject, and
\emph{all}-correct only if all three approve. \crnew{Any- and all-correct count a judgment that ends without a verdict as a rejection and majority-correct leaves it out, which matters unevenly. Such judgments are 8.5\% of Qwen3.6's, touching one in five of its variants, reach fewer than one in a hundred of V4-Flash's variants, and make up 16.5\% of R1-distill's, almost all ending without a boxed verdict rather than at the token limit, which is why R1-distill's all-correct rate falls to 36\% on the clean control. Requiring two approvals outright would lower every Qwen3.6 rate by 1.7 to 5.3 points without reordering the five conditions, leave V4-Flash unchanged, and lower R1-distill's rates by 2 to 11 points. R1-distill is also unstable on identical input, splitting 22\% of the clean controls that received three verdicts between approval and rejection, against 2\% for Qwen3.6 and 7\% for V4-Flash.}

Table~\ref{tab:error-rules} repeats the error injection experiment under
all three strategies. No strategy detects the intermediate numeric error,
whose acceptance stays within 1 point of the clean control in all six
\crnew{combinations of V4-Flash and Qwen3.6 and, for R1-distill, falls at most 1 point below it}, and in every combination \crnew{of V4-Flash and Qwen3.6} the consistent
final-answer error is accepted more often than the plain one\crnew{, while
R1-distill orders the two the other way under all three strategies}. What the
strategies change is only how far apart they place the two groups.
\crnew{For V4-Flash and Qwen3.6,} all-correct separates them furthest, but relative to any-correct it also
rejects 7.1 points more of the unedited V4-Flash controls and 18.2 points
more of the Qwen3.6 ones, so part of its strictness is disagreement
between generations of the same judge rather than error detection.

\begin{figure*}[!htbp]
  \centering
  \includegraphics[width=\textwidth]{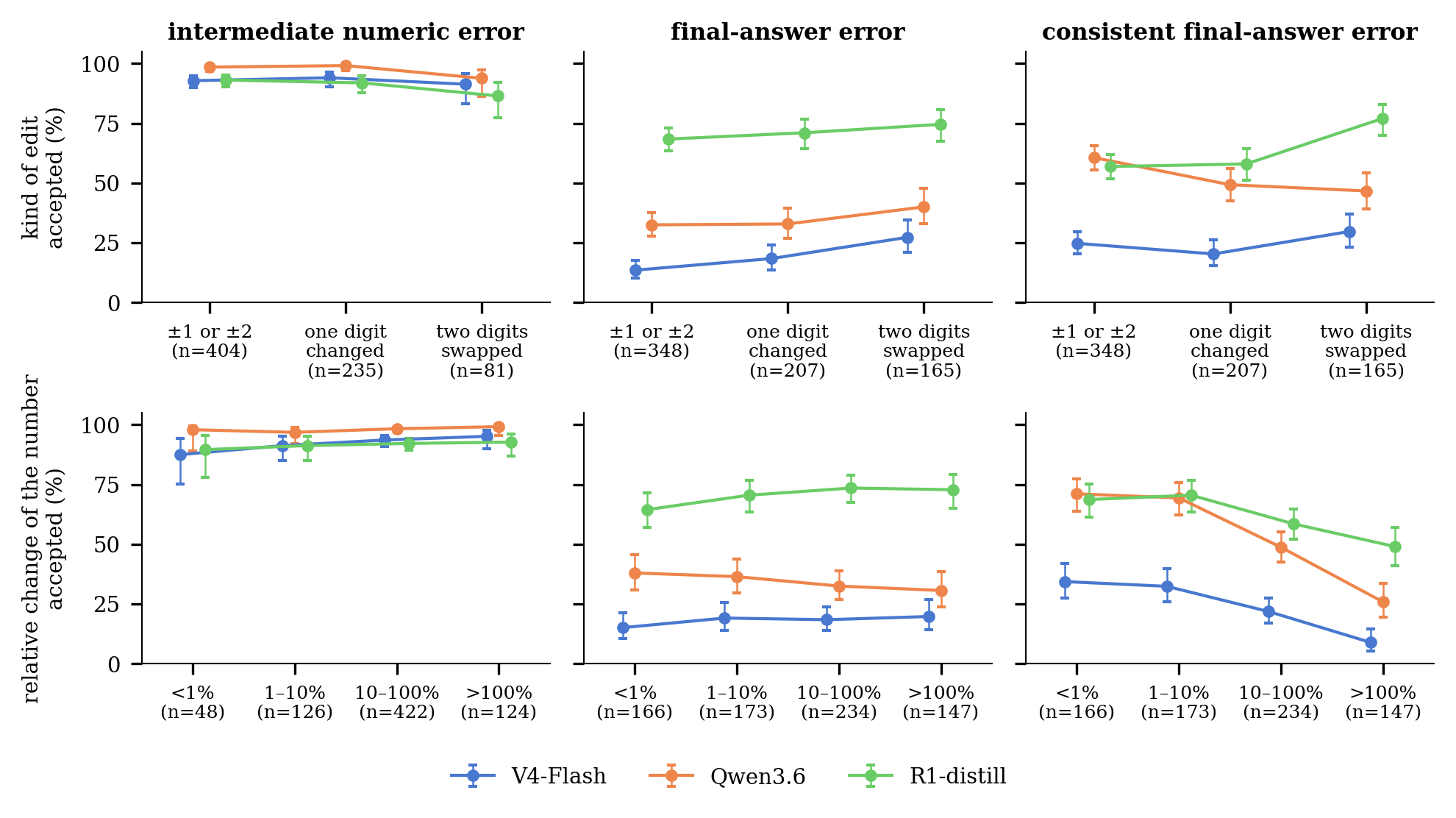}
  \caption{\crnew{Acceptance (\%, majority-correct) by the kind and the size of the injected edit,
  with 95\% Wilson intervals, for the three conditions that change a number (columns). Top, how
  the wrong number was formed; bottom, its relative change $|\text{new}-\text{old}|/|\text{old}|$.
  Each row of panels partitions the same 720 solutions of every condition, and the number of
  solutions in a category is given under its label.}}
  \label{fig:edit-size}
\end{figure*}

\paragraph{\crnew{Size and kind of the edit.}}
\crnew{Figure~\ref{fig:edit-size} splits the same 720 solutions by how the wrong number was formed and by how far it moved. The intermediate numeric error is missed whatever the edit, at 86 to 100\% acceptance in every cell of every judge, and its position between 40 and 70\% of the chain moves acceptance by at most 3 points. The plain final-answer error is also insensitive to size, within 10 points across the four size bins. Only the consistent final-answer error reacts to the size of the number. A wrong answer within 1\% of the correct one is accepted 71\% of the time by Qwen3.6, 34\% by V4-Flash and 69\% by R1-distill, one more than double the correct value 26, 9 and 49\%.}

\begin{figure*}[!htbp]
  \centering
  \includegraphics[width=\textwidth]{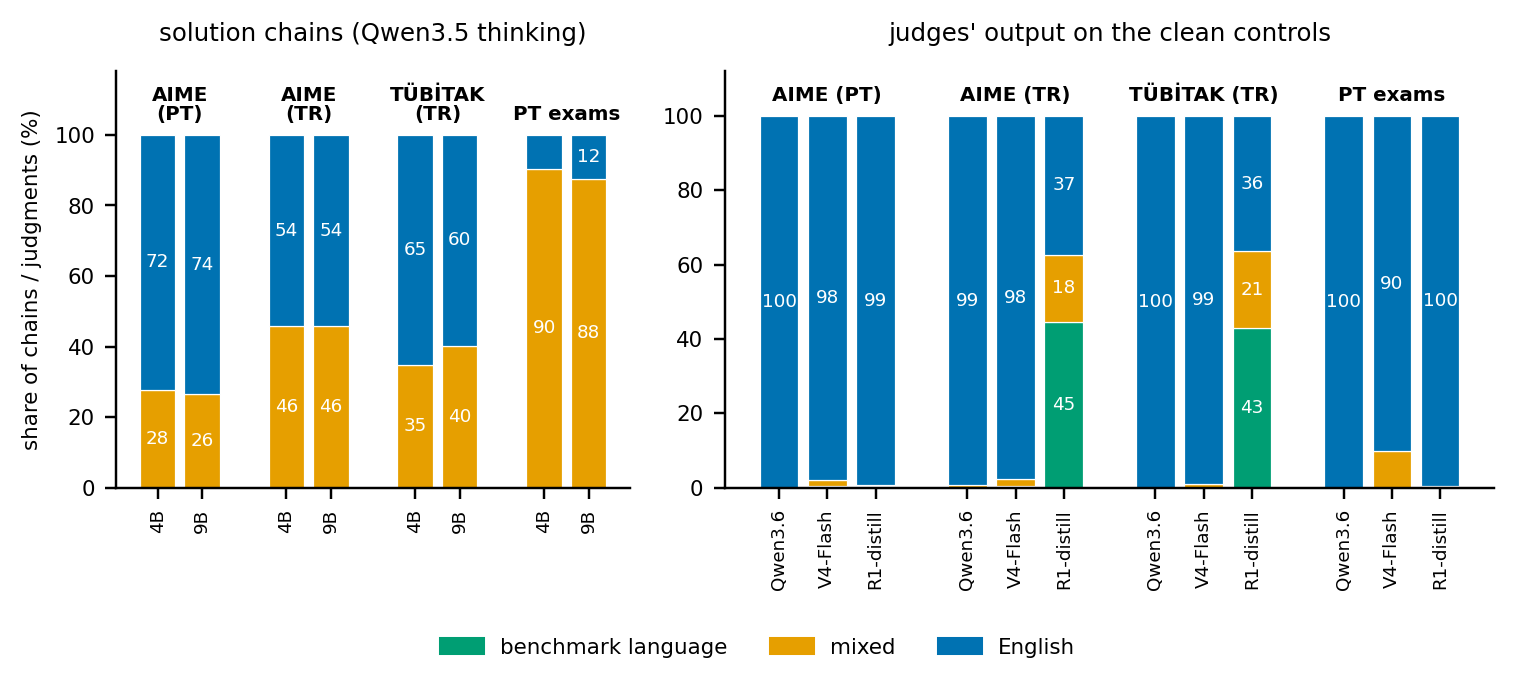}
  \caption{\crnew{Language of the error-injection study, by benchmark. Left, the 720 original solutions
  of Section~\ref{sec:method-error-injection} (Qwen3.5-4B and 9B, thinking mode), classed by the share of
  their letters, mathematics removed, in the benchmark's language (at least 80\%, \emph{benchmark
  language}), in English (at least 80\%, \emph{English}) or in between (\emph{mixed}); the edits do not
  change a chain's language, so the five variants share the classification. Right, the judges' output
  on the clean controls, thinking and visible answer together, classed the same way. On English AIME every chain and every judgment is English, so that benchmark is left out. Under the wrong-answer conditions Qwen3.6's shares stay within 1 point of these, R1-distill's all-English share on the Turkish benchmarks falls from 36 to 37\% to 19 to 26\%, and V4-Flash's mixed share on the Portuguese exams falls from 10\% to under 3\%.
  Appendix~\ref{app:language} gives the method.}}
  \label{fig:language-exp1}
\end{figure*}

\paragraph{\crnew{Rates per benchmark.}}
\crnew{Figure~\ref{fig:error-injection} gives the majority-correct rates per benchmark. The intermediate numeric error shifts no benchmark under any judge (exact McNemar on 144 solutions each, smallest $p = 0.30$), and R1-distill's loss of 15 to 27 points under a wrong final answer is significant on every benchmark ($p < 10^{-3}$). The Portuguese exams carry the shortest solutions, a median of 3.7k tokens against 12k to 24k on the other four, and every judge accepts their clean control at 99 to 100\%. The Turkish olympiad gives each judge its lowest clean control and is the one benchmark where V4-Flash accepts the consistent error less often than the plain one; R1-distill does so there and on the Portuguese exams as well (exact McNemar $p \le 1.6\times10^{-3}$ for all three reversals). On the three AIME sets, the same problems in three languages, the rates move together.}

\section{Judges, Sampling and Uncertainty}
\label{app:judges}

\crnew{Two} tables and \crnew{six} figures support Section~\ref{sec:results-coincide}\crnew{, one
group per subsection}.

\subsection{\crnew{The Judge Matrix}}
\label{app:judge-matrix}

Table~\ref{tab:judge-matrix-earlier} gives the difference Pass@$64$ $-$
CoT-Pass@$64$ for every earlier-generation solver under all four judges,
and Table~\ref{tab:judge-matrix-current} for every current-generation
solver under the two judges that cover both generations: together the
per-model form of the claim that the collapse does not depend on the
judge.

The verdict threshold of Section~\ref{sec:method-comparison} removes
from the comparison grid the Gemma-4 base checkpoints (too
few correct answers to support a rate), Qwen3.5-0.8B (under the threshold on three of the four benchmarks \crnew{as a base checkpoint and on all four in non-thinking mode}) and Qwen3.5-2B (over it only
as a base checkpoint).

\begin{table}[!htbp]
\centering
\footnotesize
\setlength{\tabcolsep}{3pt}
\begin{tabular}{lrrrr}
\toprule
Solver & Qwen3.6 & Gemma4 & V4-Flash & R1-distill \\
\midrule
Q2.5-7B & 15.6 & 13.9 & 18.0 & 15.6 \\
Q2.5-32B & 20.5 & 22.1 & 23.0 & 21.3 \\
Q2.5-7B-I & 22.1 & 23.0 & 23.8 & 22.1 \\
Q2.5-32B-I & 20.5 & 20.5 & 21.3 & 21.3 \\
\midrule
Average & 19.7 & 19.9 & 21.5 & 20.1 \\
\bottomrule
\end{tabular}
\caption{Pass@64 $-$ CoT-Pass@64 per earlier-generation solver (rows),
averaged over the four 64-sample benchmarks (weighted by how many
questions each contributes, as in Figures~\ref{fig:k-ladder}
and~\ref{fig:bootstrap}) under all four judges
(columns). Judge generation budgets: 16{,}384 for Qwen3.6, Gemma4 and
R1-distill, 20{,}480 for V4-Flash; every judge runs in thinking mode with
three judgments per solution. The last row averages the solvers.
Q2.5 = Qwen2.5 (-I Instruct).}
\label{tab:judge-matrix-earlier}
\end{table}
\begin{table}[!htbp]
\centering
\footnotesize
\setlength{\tabcolsep}{3pt}
\begin{tabular}{lrr}
\toprule
Solver & Qwen3.6 & Gemma4 \\
\midrule
Q3.5-4B-B & 0.8 & 1.6 \\
Q3.5-9B-B & 0.0 & 0.8 \\
Q3.5-4B & 4.9 & 6.6 \\
Q3.5-9B & 3.3 & 4.1 \\
G4-E2B & 8.2 & 8.2 \\
G4-E4B & 7.4 & 4.9 \\
\midrule
Average & 4.1 & 4.4 \\
\bottomrule
\end{tabular}
\caption{As Table~\ref{tab:judge-matrix-earlier}, for the
current-generation solvers under the two judges that cover them.
Q3.5 = Qwen3.5 (-B base), G4 = Gemma-4-it.}
\label{tab:judge-matrix-current}
\end{table}

\subsection{\crnew{The $k$ Ladder and Question-Level Uncertainty}}
\label{app:k-ladder}

\begin{figure}[!htbp]
  \centering
  \includegraphics[width=\columnwidth]{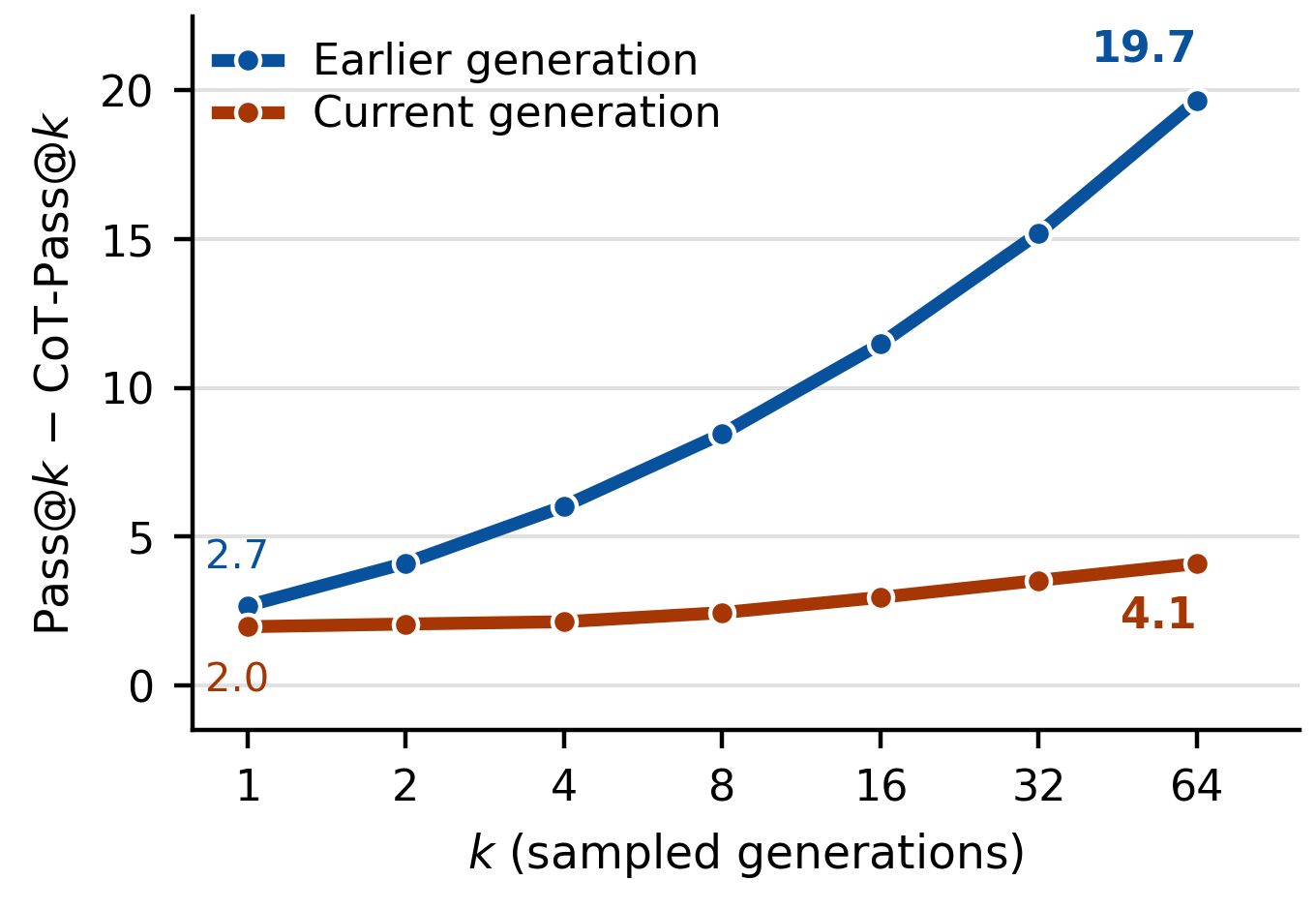}
  \caption{The difference Pass@$k$ $-$ CoT-Pass@$k$ against the number of
  sampled generations, under the anchor judge Qwen3.6 at 16k. Each solver
  is a weighted mean over its four benchmarks, weighted by how many
  questions each contributes; each curve is then the unweighted mean over
  the solvers of that generation. Endpoints are labelled;
  Figure~\ref{fig:bootstrap} gives the per-solver spread at $k=64$.}
  \label{fig:k-ladder}
\end{figure}

\begin{figure}[!htbp]
  \centering
  \includegraphics[width=\columnwidth]{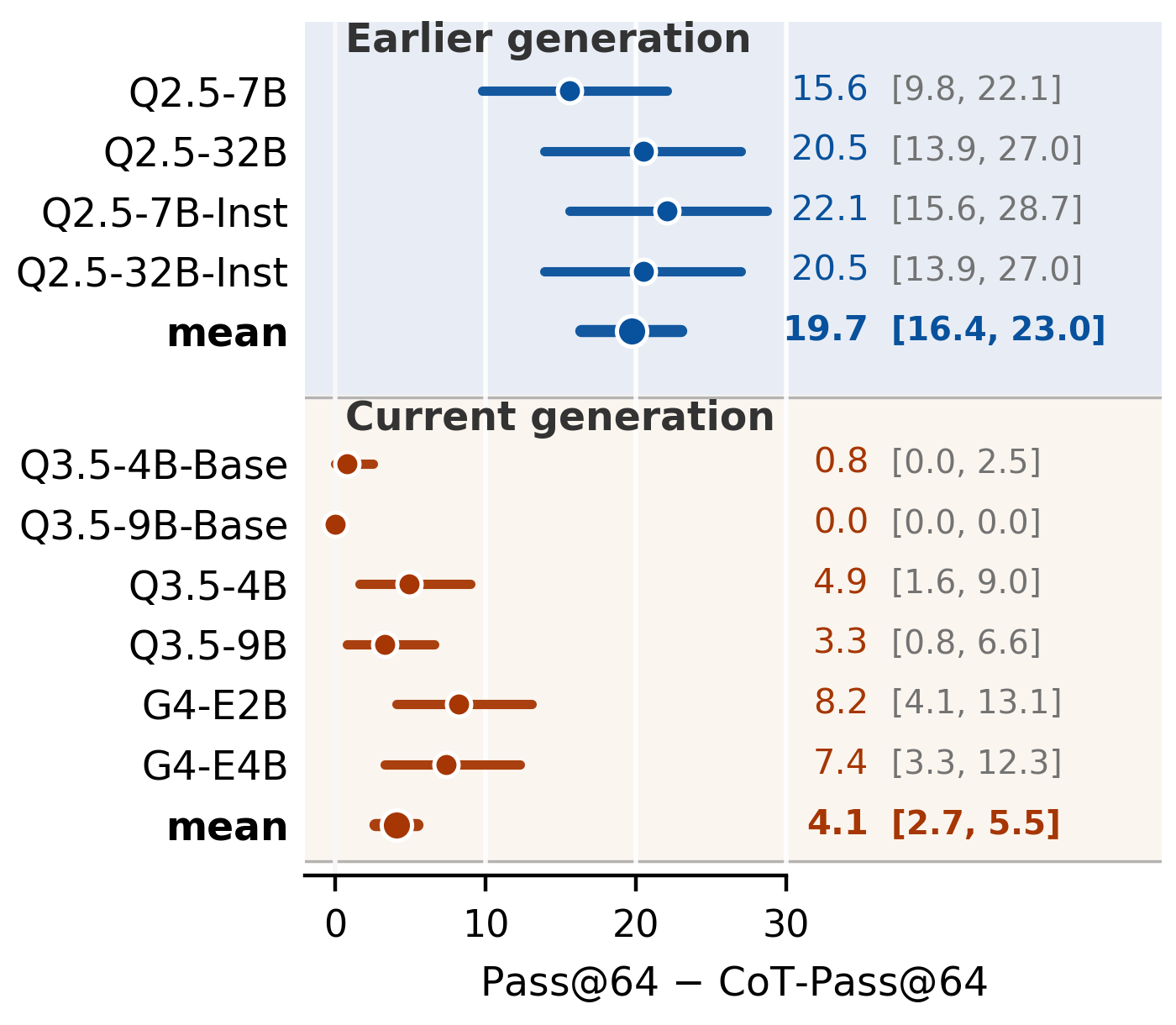}
  \caption{Question-level bootstrap 95\% intervals for Pass@$64$ $-$
  CoT-Pass@$64$ under the anchor judge Qwen3.6 at 16k, in points, with the
  numeric interval at the right of each row. Each solver is a weighted
  mean over its four benchmarks, weighted by how many questions each
  contributes; \emph{mean} is the unweighted mean over the solvers of a
  generation. 10{,}000 resamples of questions, percentile intervals; cells
  are resampled independently. Q2.5/Q3.5 = Qwen2.5/Qwen3.5, G4 =
  Gemma-4-it.}
  \label{fig:bootstrap}
\end{figure}

Figure~\ref{fig:k-ladder} follows the same difference as the number of
sampled generations grows. Figure~\ref{fig:bootstrap} reports the
question-level bootstrap intervals quoted in the text, per solver.

One row of Figure~\ref{fig:bootstrap} needs a warning. Qwen3.5-9B-Base has
a difference of exactly zero on each of the four benchmarks, so every
resample of its questions returns zero and its interval collapses to a
point. That is a structural limit of resampling an all-zero indicator, not
a statement that the value is known without uncertainty, and the same
applies to any cell-level interval of the form $[0,0]$.

\subsection{\crnew{Acceptance of Correct Generations}}
\label{app:acceptance}

\begin{figure*}[!htbp]
  \centering
  \includegraphics[width=\textwidth]{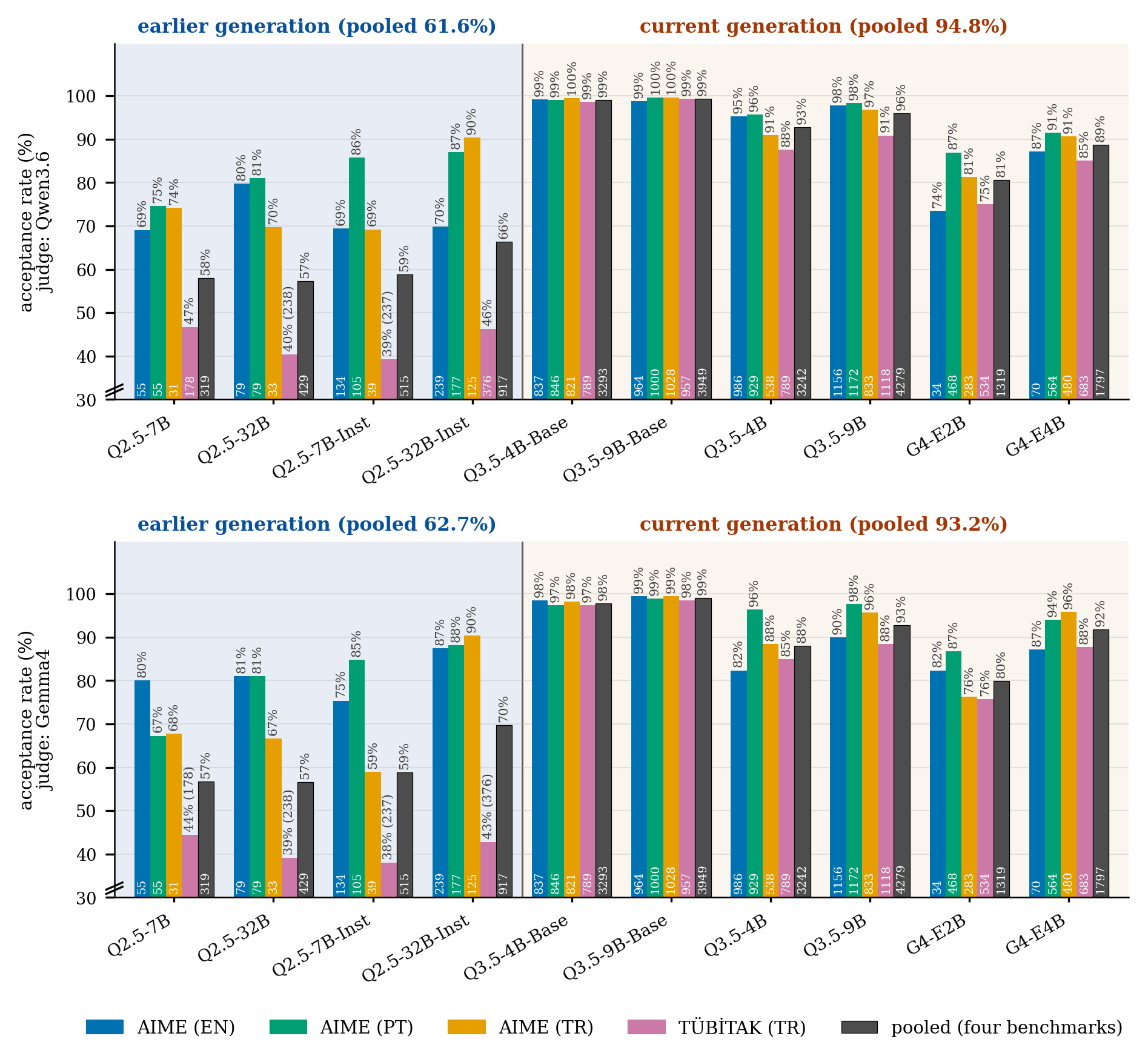}
  \caption{\crnew{Acceptance rate of answer-correct generations per solver and benchmark, the
  share of a solver's correct generations whose reasoning chain the judge approved, under
  Qwen3.6 (top) and Gemma4 (bottom), for the solvers of
  Section~\ref{sec:results-coincide} at the 16k budget. The number above each bar is that
  rate in percent; the number at its base is the count of answer-correct generations in
  the cell, given in parentheses after the rate where the bar is too short. The rejected
  share is 100 minus the rate. The dark grey bar of each solver gives its rate pooled over
  the four benchmarks, with the pooled count at its base, and the panel headers give the
  rate pooled over each generation. The vertical axis starts at 30.
  Q2.5/Q3.5 = Qwen2.5/Qwen3.5 (-Inst instruct, -Base base), G4 = Gemma-4-it.}}
  \label{fig:acceptance}
\end{figure*}

\crnew{Section~\ref{sec:results-coincide} reports the collapse as the difference
Pass@$64$ $-$ CoT-Pass@$64$, and at $k{=}64$ that difference can shrink for a
reason that has nothing to do with the judge. A question keeps its CoT-Pass@$64$
as long as one of its correct generations is approved, so where a question holds
many correct generations the judge can reject most of them without moving the
metric. Two figures, here and in Appendix~\ref{app:saturation}, separate that
saturation from the verdicts.
Figure~\ref{fig:acceptance} gives the quantity the difference hides, the
acceptance rate of answer-correct generations, per solver and benchmark and under
both judges. The two generations never overlap, 57 to 70\% against 80 to 99\% per solver, and on the native Turkish olympiad the anchor judge approves only 39 to 47\% of the earlier generation's correct generations and 75 to 99\% of the current generation's. Averaged over the ten solvers the Turkish translation sits within 3 points of the English original under both judges (Wilcoxon \crnew{signed-rank test~\citep{wilcoxon1945individual}}, $p = 0.49$), the Portuguese translation runs 6 points above it under the anchor judge ($p = 0.004$) and 3 above under Gemma4 (\crnew{$p = 0.25$}), and the olympiad sits 15 and 14 points below the Turkish translation ($p = 0.002$), so the language of the problem does not lower the rate and the benchmark does.}

\subsection{\crnew{Saturation and Conditioning on $c$}}
\label{app:saturation}

\begin{figure*}[!htbp]
  \centering
  \includegraphics[width=\textwidth]{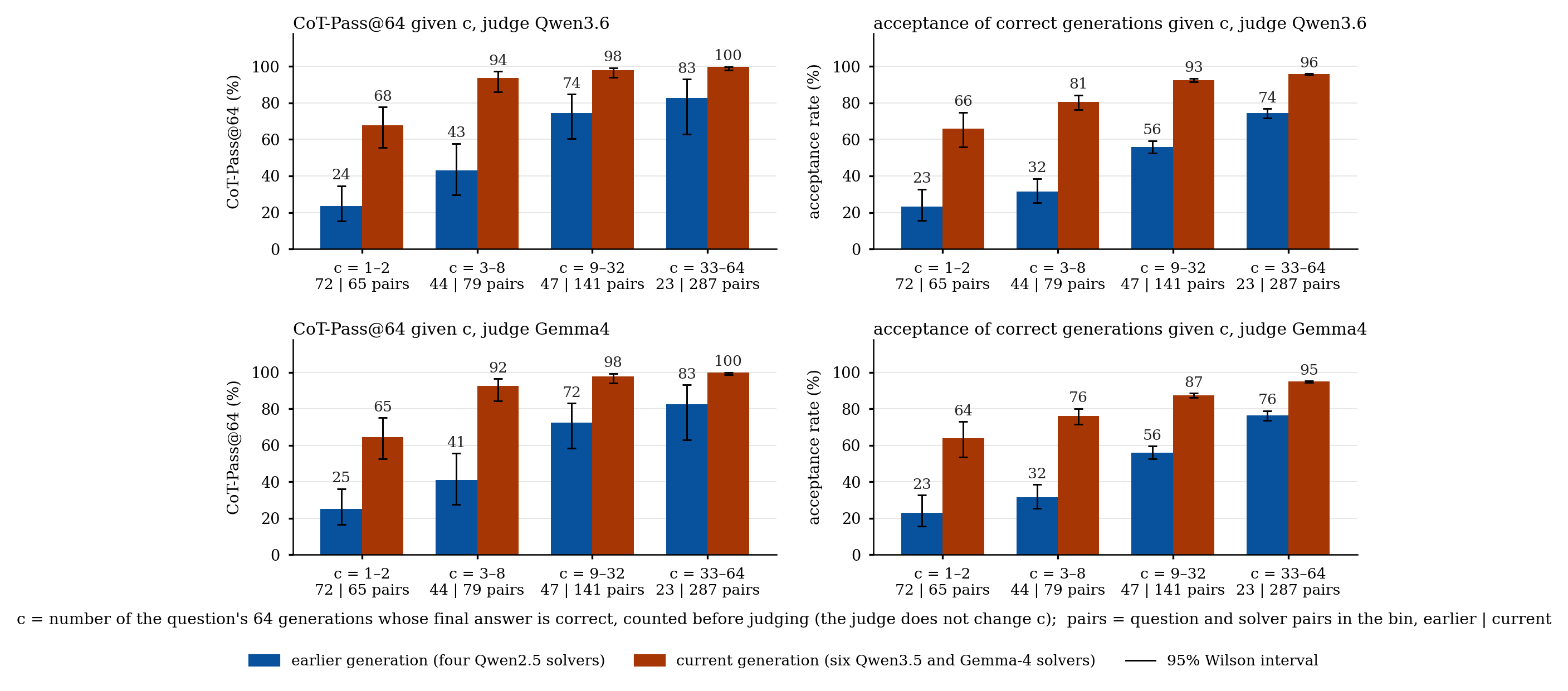}
  \caption{\crnew{CoT-Pass@$64$ after conditioning on $c$, the number of a question's 64
  generations whose final answer is correct, counted before judging, for the ten solvers of
  Section~\ref{sec:results-coincide} at the 16k budget, pooled over the four benchmarks,
  under both judges (rows), with 95\% Wilson intervals. Left, the share of question and
  solver pairs in the bin in which the judge approved at least one correct generation,
  which is CoT-Pass@$64$ for that bin since Pass@$64$ there is 100 by construction;
  right, the share of answer-correct generations in the bin that the judge approved, the
  acceptance rate of Figure~\ref{fig:acceptance} at fixed $c$, so that the two generations
  can be compared at the same $c$. The tick labels give the number of question and solver
  pairs in each bin, earlier | current. Per benchmark, the Turkish olympiad has the lowest value in both panels in
  every bin of the earlier generation, and Figure~\ref{fig:acceptance} gives the
  acceptance rates per benchmark.}}
  \label{fig:saturation}
\end{figure*}

\crnew{The tick labels of Figure~\ref{fig:saturation} give the number of question and
solver pairs in each bin of $c$. The earlier generation's solved questions sit
at $c \le 8$, where 80 of 116 pairs lose every correct generation, and the current
generation's at $c \ge 33$, where 1 of 287 does, so the saturation is real and
it favours the current generation. The figure then holds $c$
fixed. At every $c$ the current generation keeps more of its questions and more
of its correct generations. At $c = 1$--$2$ CoT-Pass@$64$ is 24 against 68 under
the anchor judge and the judge accepts 23\% against 66\% of the correct
generations; at $c = 33$--$64$ it is 83 against 100 and 74\% against 96\%, so in both
generations the judges reject most where a question holds few correct
generations. The gap between the generations therefore survives conditioning on $c$, and what
saturation explains is only how the current generation's remaining rejections,
4 to 34\% of its correct generations, turn into a difference of zero.}

\subsection{\crnew{Language of the Chains}}
\label{app:language}

\crnew{Solvers are prompted in the benchmark's language and their chains may switch
into English (Section~\ref{sec:method-setup}). To measure how often, every answer-correct
generation of the judge matrix is stripped of its mathematics (display and inline
formulas, boxed answers, LaTeX commands, digits and operator symbols) and cut into
lines and sentences of at least twenty letters; each segment receives one label from
GlotLID \citep{kargaran2023glotlid}, and segments that restate the question are dropped.
A chain is \emph{target} when at least 80\% of its letters are in the benchmark's language,
\emph{English} when at least 80\% are English, and \emph{mixed} otherwise. As a second layer,
MaskLID \citep{kargaran2024masklid}, which detects code switching by iterative masking, is
run on the first 1{,}000 words of every earlier-generation chain, of a fifth of the current
generation's chains and of every panel chain; the two layers agree on whether a chain is
mixed for 93\% of the chains they share. Labels outside the three languages cover under 3\%
of the letters, and a threshold of 70 or 90\% changes the \emph{target} share by at most one
point.}

\begin{figure*}[!htbp]
  \centering
  \includegraphics[width=\textwidth]{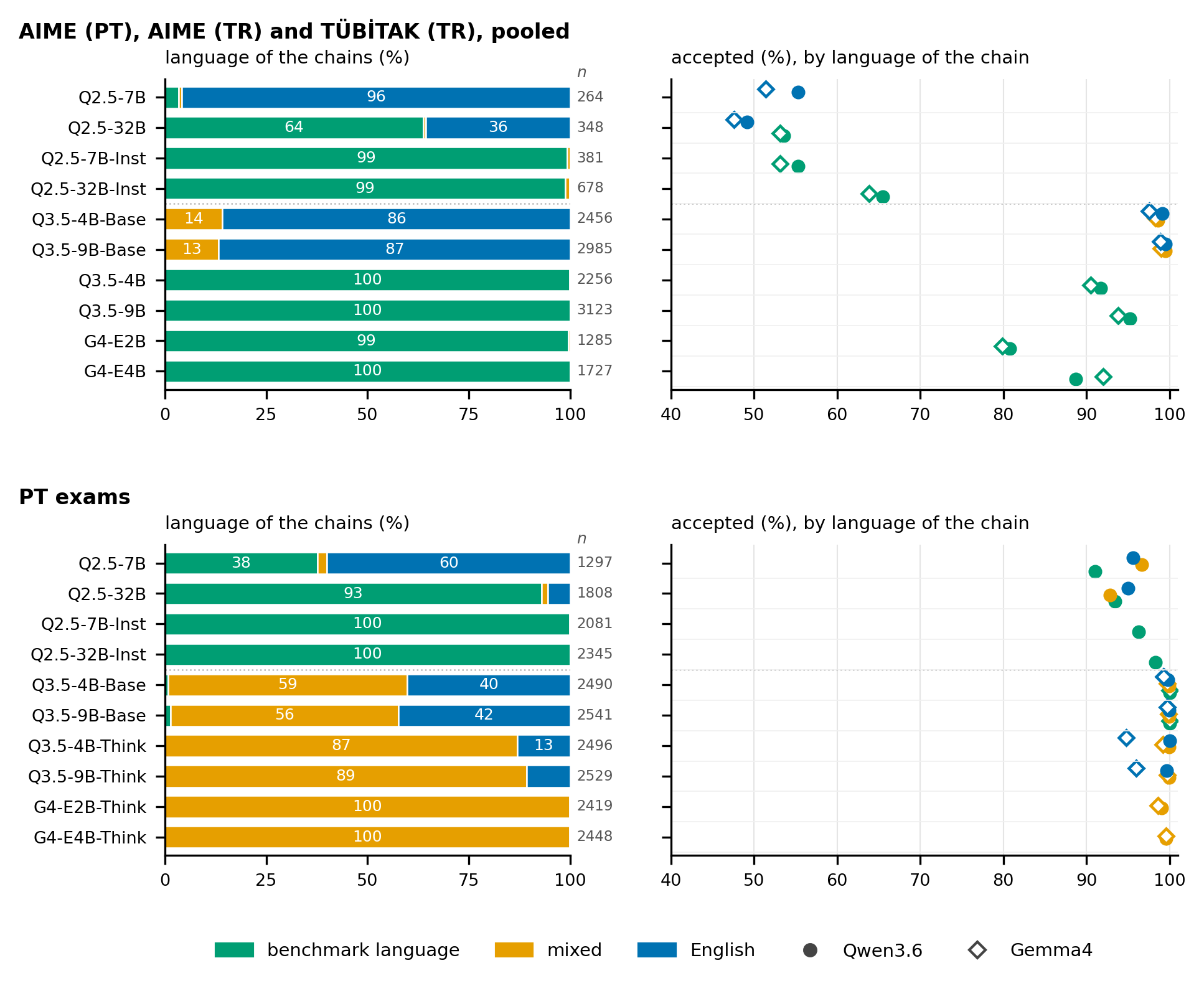}
  \caption{\crnew{Language of the reasoning chains (left) and acceptance by language (right) for the answer-correct generations of the ten solvers of Section~\ref{sec:results-coincide} at the 16k budget, on the three non-English benchmarks pooled (top) and on the Portuguese exams (bottom), where thinking runs stand in for the non-thinking runs they lack. A chain is in the \emph{benchmark language} when at least 80\% of its letters, mathematics removed, are in that language, \emph{English} when at least 80\% are English, \emph{mixed} otherwise. Acceptance is the majority verdict of Qwen3.6 (filled) and Gemma4 (open); a class with fewer than 20 chains of a solver is not plotted, and $n$ is the number of chains. The dotted line separates the earlier generation from the current one. Pooling mixes benchmarks with different acceptance rates, so within-solver comparisons are read per benchmark in Appendix~\ref{app:language}. Q2.5/Q3.5 = Qwen2.5/Qwen3.5 (-Inst instruct, -Base base, -Think thinking mode), G4 = Gemma-4-it.}}
  \label{fig:code-switching}
\end{figure*}

\begin{figure*}[!htbp]
  \centering
  \includegraphics[width=\textwidth]{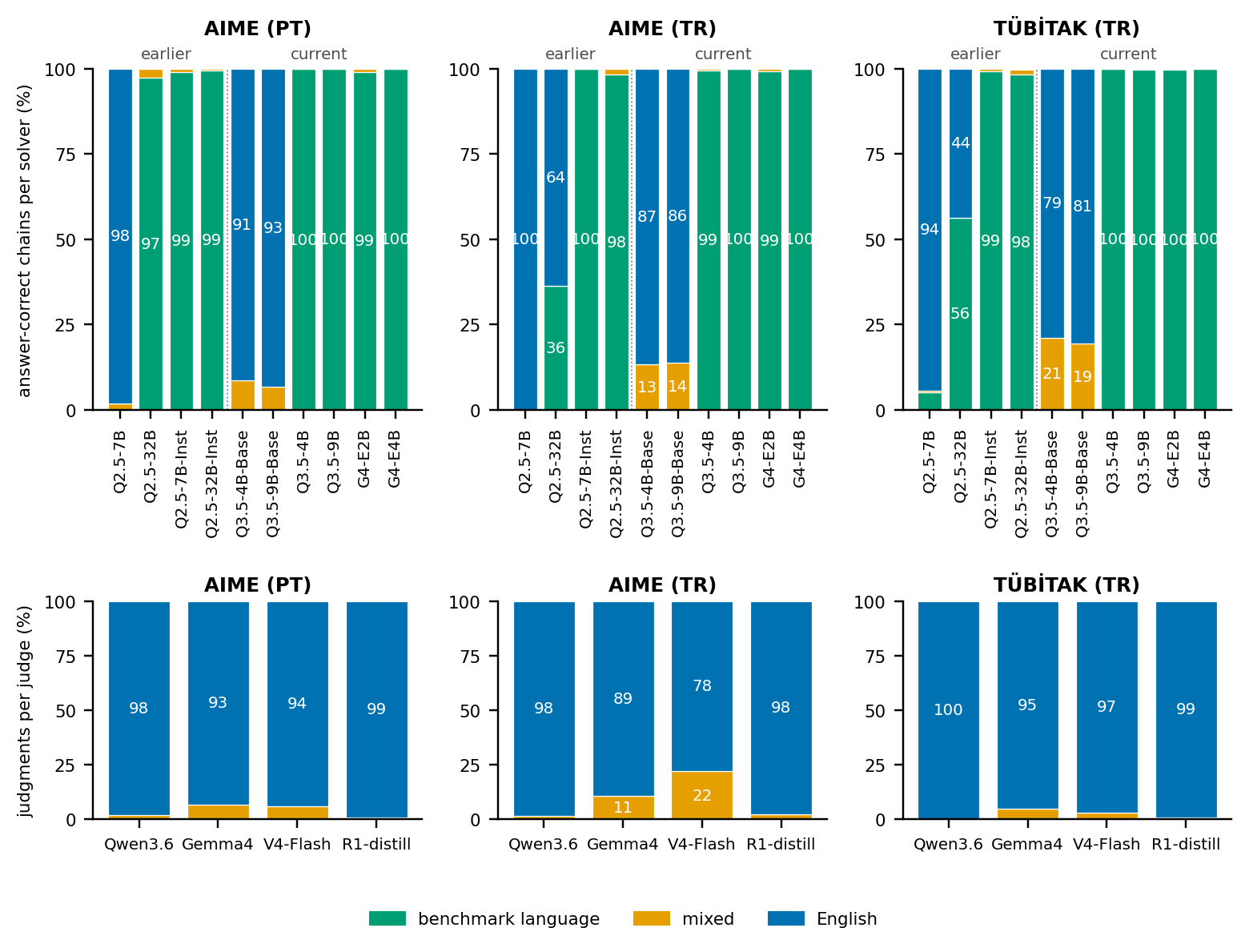}
  \caption{\crnew{Language of the answer-correct chains and of the judgments in the judge matrix of
  Section~\ref{sec:results-coincide}, per benchmark. Top, the ten solvers at the 16k budget, earlier
  generation left of the dotted line and current generation right of it (-Inst instruct, -Base base); bottom,
  a sample of 300 judgments per judge and benchmark. Classes as in Figure~\ref{fig:language-exp1}.
  English AIME is left out, every chain and judgment there being English.
  Figure~\ref{fig:code-switching} gives the same shares pooled over the three benchmarks together with
  acceptance by class.}}
  \label{fig:language-exp2}
\end{figure*}

\crnew{The language of a chain is a property of the solver
(Figures~\ref{fig:code-switching} and~\ref{fig:language-exp2}). The Qwen3.5 base
checkpoints, which think by default, write 79 to 93\% of their chains on the
non-English benchmarks in English and the rest mixed, while the instruction-tuned solvers
of both generations answer in the benchmark's language at least 97\% of the time; only the
Qwen2.5 base checkpoints split, and the Portuguese exams show the same division. Because
the classes coincide with solvers, acceptance by language cannot be read across solvers.
Within a solver the judges accept English and mixed chains alike, within 3 points for every
Qwen3.5 base cell, and the one solver with enough chains of both kinds, Qwen2.5-32B on the
Turkish olympiad, is accepted 48\% of the time in English and 35\% in Turkish.}

\crnew{The error-injection panel holds no chain in the benchmark's language
(Figure~\ref{fig:language-exp1}). There the verdicts move with the language under two
judges on the Turkish benchmarks. On the Turkish olympiad V4-Flash accepts 94\% of the
mixed clean chains against 69\% of the all-English ones and R1-distill 85 against 74\%,
and R1-distill also accepts the wrong final answer more often in a mixed chain, 72 against
49\% there and 73 against 63\% on the Turkish AIME set; Qwen3.6 stays within 4 points on
every clean control, and on the Portuguese benchmarks every judge's clean controls stay within 7.}

\crnew{The judges themselves reason in English. On a sample of 300 judgments per judge and
benchmark from the judge matrix, all four judges write 78 to 100\% of their judgments on the
Turkish and Portuguese benchmarks entirely in English and none in the benchmark's language,
whatever the language of the chain (Figure~\ref{fig:language-exp2}). The error-injection
panel gives the same picture with one exception. On the two Turkish benchmarks R1-distill
thinks mostly in Turkish in 40 to 47\% of its judgments while its visible verdict stays in
English, and the judgments it thinks through in Turkish approve more often, 78 against 64\%
of the clean controls on the Turkish AIME set and 65 against 62\% on the olympiad.}

\section{\crnew{Token Budgets and Generation Mode}}
\label{app:conditions}

\subsection{\crnew{Stops at the Max-Token Limit}}
\label{app:stops}

\crnew{Table~\ref{tab:conditions} supports Section~\ref{sec:results-conditions} with the share of
solver generations and of judgments that stop at the max-token limit, per benchmark and
solver, with the difference each judge reports.}

\begin{table}[!htbp]
\centering
\footnotesize
\setlength{\tabcolsep}{2pt}
\begin{tabular}{llrrrrr}
\toprule
 & & \multicolumn{1}{c}{solver} & \multicolumn{2}{c}{Gemma4} &
\multicolumn{2}{c}{Qwen3.6} \\
\cmidrule(lr){4-5}\cmidrule(lr){6-7}
Benchmark & Solver & \multicolumn{1}{c}{stops} & stops &
\multicolumn{1}{c}{diff.} & stops & \multicolumn{1}{c}{diff.} \\
\midrule
AIME (EN) & Q3.5-4B & 89.6 & 35.0 & 3.33 & 15.7 & 0.00 \\
 & Q3.5-9B & 83.9 & 33.0 & 3.33 & 17.5 & 0.00 \\
 & G4-E2B & 0.2 & 2.1 & 6.67 & 6.5 & 3.33 \\
 & G4-E4B & 1.4 & 2.7 & 10.00 & 6.0 & 3.33 \\
\addlinespace[2pt]
AIME (PT) & Q3.5-4B & 47.7 & 53.5 & 13.33 & 5.2 & 0.00 \\
 & Q3.5-9B & 44.0 & 49.8 & 0.00 & 3.6 & 0.00 \\
 & G4-E2B & 0.1 & 9.8 & 3.33 & 1.4 & 3.33 \\
 & G4-E4B & 0.2 & 7.5 & 6.67 & 1.9 & 0.00 \\
\addlinespace[2pt]
AIME (TR) & Q3.5-4B & 44.2 & 51.0 & 0.00 & 10.0 & 0.00 \\
 & Q3.5-9B & 35.6 & 53.8 & 10.00 & 8.3 & 0.00 \\
 & G4-E2B & 0.5 & 7.6 & 3.33 & 3.0 & 3.33 \\
 & G4-E4B & 0.2 & 9.8 & 6.67 & 3.2 & 0.00 \\
\addlinespace[2pt]
TÜB\.ITAK (TR) & Q3.5-4B & 44.4 & 47.6 & 6.25 & 3.6 & 0.00 \\
 & Q3.5-9B & 35.6 & 49.1 & 6.25 & 3.4 & 0.00 \\
 & G4-E2B & 0.8 & 13.6 & 3.12 & 0.2 & 0.00 \\
 & G4-E4B & 0.0 & 10.3 & 9.38 & 0.1 & 9.38 \\
\addlinespace[2pt]
PT exams & Q3.5-4B & 3.6 & 5.0 & 0.00 & 1.5 & 0.00 \\
 & Q3.5-9B & 1.7 & 3.5 & 0.00 & 1.3 & 0.00 \\
 & G4-E2B & 0.1 & 0.7 & 0.00 & 0.2 & 0.60 \\
 & G4-E4B & 0.0 & 0.4 & 0.00 & 0.2 & 0.00 \\
\bottomrule
\end{tabular}
\caption{\crnew{Stops at the max-token limit on both sides,} in thinking mode at the 16k generation
budget: the share of solver generations that stop at the max-token limit
(stops, \%), and per judge the share of its judgments that do, with the
difference Pass@$k_{\max}$ $-$ CoT-Pass@$k_{\max}$ (diff.) under that
judge. $k_{\max}$ is 64, or 16 for the Portuguese exams; stop rates come
from each generation's recorded finish state. Q3.5 = Qwen3.5,
G4 = Gemma-4-it.}
\label{tab:conditions}
\end{table}

\subsection{Judge Budgets and the Difference}
\label{app:capping}

\begin{figure}[!htbp]
  \centering
  \includegraphics[width=\columnwidth]{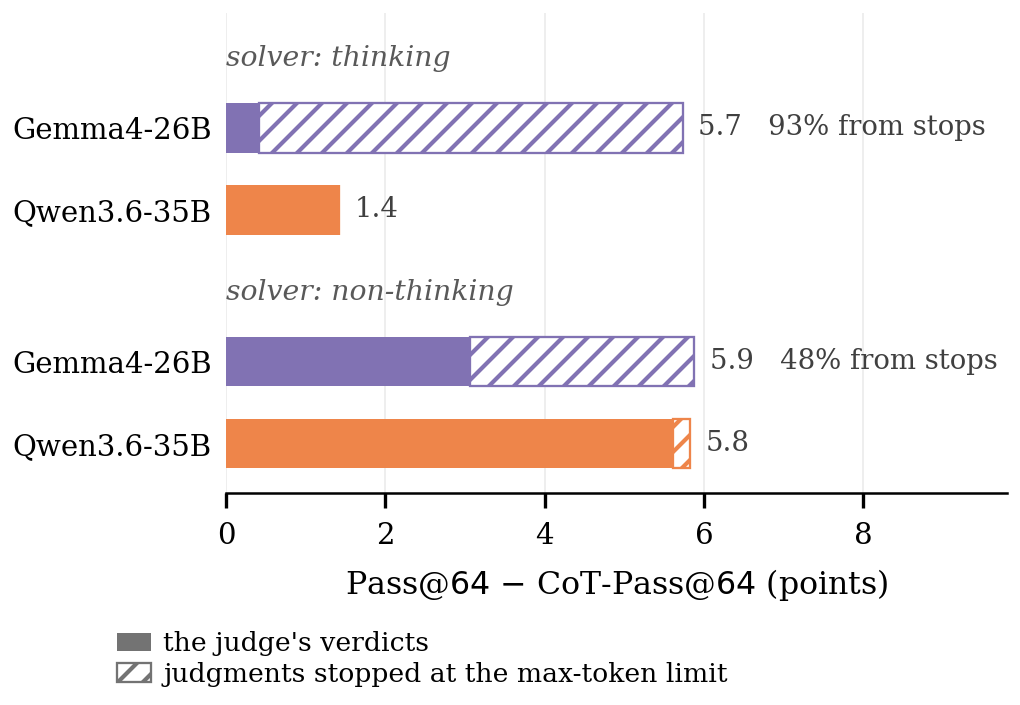}
  \caption{The difference Pass@$64$ $-$ CoT-Pass@$64$ per judge and solver
  mode, split into the part carried by the judge's verdicts (solid) and
  the part that disappears when judgments stopped at the max-token limit
  are counted as approvals (hatched). Cell means over the four 64-sample
  benchmarks and the four current-generation solvers, 16k budget on both
  sides.}
  \label{fig:capping}
\end{figure}

Figure~\ref{fig:capping} decomposes the difference Pass@$64$ $-$
CoT-Pass@$64$ that each judge reports, under the counterfactual of
Section~\ref{sec:method-budget}: judgments that stop at the max-token
limit are counted as approvals instead, and what survives is \crnew{a lower bound on}
the part carried by the judge's verdicts\crnew{, since a stopped judgment that would
have rejected is credited as an approval}. The scope is the fully crossed design of
Section~\ref{sec:results-conditions}: the four current-generation
solvers on the four 64-sample benchmarks, in both generation modes, every
cell scored by both judges at the 16k budget.

The same counterfactual applied to the earlier generation certifies the
other side of the contrast: the four Qwen2.5 solvers on the same four
benchmarks, every correct solution judged, at the same 16k solver
budget. Crediting every stopped judgment as an approval leaves, of the
average difference each judge reports in
Table~\ref{tab:judge-matrix-earlier}, \crnew{at least} 15.2 of the anchor judge's 19.7
points, 12.5 of Gemma4's 19.9, 18.6 of R1-distill's 20.1, and 20.1 of
V4-Flash's 21.5. Under the two judges that rarely stop at their own
limit the difference barely moves: the earlier generation's side of the
generational contrast is carried by verdicts, not by the judges'
budgets.

\subsection{The Budget Ladder}
\label{app:ladder}

Table~\ref{tab:ladder} supports
Section~\ref{sec:results-conditions}: the two thinking-mode solvers of
Section~\ref{sec:method-error-injection} at their low and raised
generation budgets, row by row. \crnew{The ladder is reported under the anchor judge alone: at the raised budgets Gemma4 stopped at its own max-token limit so often that serving slowed and its runs could not be completed, the instrument buckling under the budgets it was given.}

\begin{table}[!htbp]
\centering
\footnotesize
\setlength{\tabcolsep}{2pt}
\begin{tabular}{llrrrrrr}
\toprule
 & & \multicolumn{2}{c}{stops at max} & \multicolumn{2}{c}{Pass@1} & \multicolumn{2}{c}{Pass@64} \\
 & & \multicolumn{2}{c}{tokens (\%)} & \multicolumn{2}{c}{(\%)} & \multicolumn{2}{c}{(\%)} \\
\cmidrule(lr){3-4}\cmidrule(lr){5-6}\cmidrule(lr){7-8}
Benchmark & Solver & low & high & low & high & low & high \\
\midrule
AIME (EN) & 4B & 89.6 & 18.4 & 10.5 & 75.4 & 40.0 & 96.7 \\
 & 9B & 83.9 & 8.9 & 16.2 & 84.5 & 40.0 & 96.7 \\
\addlinespace[2pt]
AIME (PT) & 4B & 47.7 & 7.6 & 51.0 & 71.0 & 83.3 & 93.3 \\
 & 9B & 44.0 & 3.8 & 55.3 & 78.2 & 80.0 & 96.7 \\
\addlinespace[2pt]
AIME (TR) & 4B & 44.2 & 5.3 & 51.8 & 68.5 & 80.0 & 93.3 \\
 & 9B & 35.6 & 2.9 & 62.6 & 76.0 & 90.0 & 93.3 \\
\addlinespace[2pt]
TÜB\.ITAK (TR) & 4B & 44.4 & 6.0 & 45.4 & 65.5 & 87.5 & 93.8 \\
 & 9B & 35.6 & 3.0 & 55.4 & 75.4 & 90.6 & 93.8 \\
\bottomrule
\end{tabular}
\caption{The budget ladder: the same solver on the same benchmark at its
low and high generation budget. Low is 16{,}384 tokens everywhere; high is
32{,}768, except 65{,}536 on English AIME (the text explains the jump).
Share of solver generations stopping at the max-token limit, Pass@$1$ and
Pass@$64$; thinking mode throughout. 4B/9B = Qwen3.5.}
\label{tab:ladder}
\end{table}

\subsection{Generation Mode and Language}
\label{app:mode}

Table~\ref{tab:language} supports
Section~\ref{sec:results-conditions}: the same thirty AIME problems in
three languages, solved by the same two solvers in both generation modes.
Neither reading of the language difference is more correct than the
other: both are conditional on a mode and a budget a paper would
ordinarily report in a footnote.

\begin{table}[!htbp]
\centering
\footnotesize
\setlength{\tabcolsep}{3pt}
\begin{tabular}{llrrr}
\toprule
Mode & Lang. & \multicolumn{1}{c}{stops at max} & \multicolumn{1}{c}{Pass@1} & \multicolumn{1}{c}{Pass@64} \\
 & & \multicolumn{1}{c}{tokens (\%)} & \multicolumn{1}{c}{(\%)} & \multicolumn{1}{c}{(\%)} \\
\midrule
non-thinking & EN & 27.9 & 55.8 & 91.7 \\
 & PT & 2.4 & 54.7 & 95.0 \\
 & TR & 16.6 & 35.7 & 88.3 \\
\addlinespace[2pt]
thinking & EN & 86.7 & 13.4 & 40.0 \\
 & PT & 45.8 & 53.2 & 81.7 \\
 & TR & 39.9 & 57.2 & 85.0 \\
\bottomrule
\end{tabular}
\caption{The same thirty AIME problems in English and in the Portuguese
and Turkish translations: Qwen3.5-\{4B,9B\} at the 16k generation budget,
both modes. Columns give the share of generations that stop at the
max-token limit, Pass@1 and Pass@64, averaged over questions and over the
two models.}
\label{tab:language}
\end{table}

\begin{figure*}[!htbp]
  \centering
  \includegraphics[width=\textwidth]{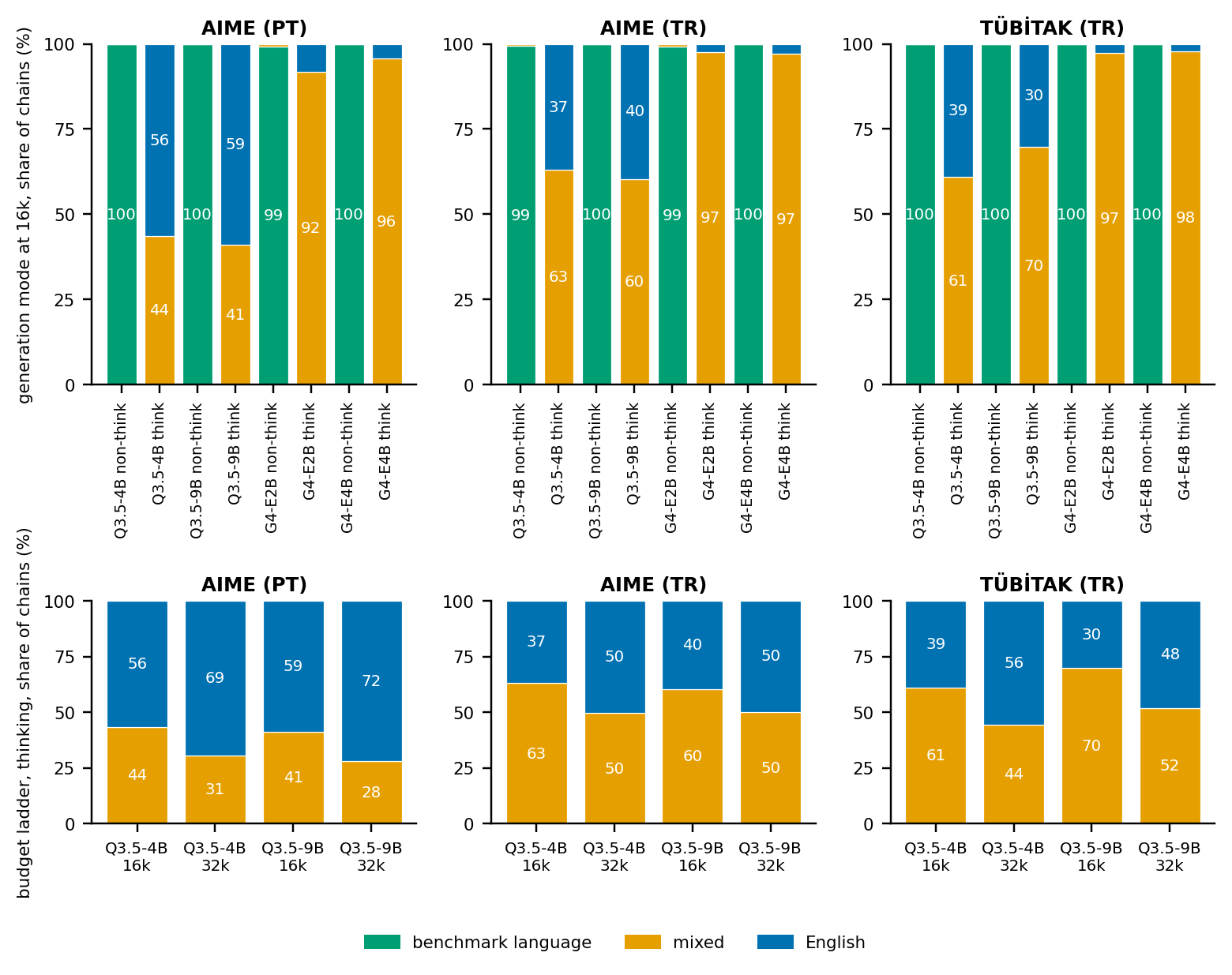}
  \caption{\crnew{Language of the answer-correct chains under the generation mode and the token budget of
  Section~\ref{sec:results-conditions}, per benchmark. Top, the four current-generation solvers at the
  16k budget in non-thinking and in thinking mode; bottom, the two Qwen3.5 thinking-mode solvers at the low
  and the raised generation budget of the ladder (Table~\ref{tab:ladder}). Classes as in
  Figure~\ref{fig:language-exp1}; English AIME left out. Turning thinking on moves every chain out of
  the benchmark's language, and raising the budget moves the mixed chains further toward English.}}
  \label{fig:language-exp3}
\end{figure*}

\end{document}